\documentclass{article}

\PassOptionsToPackage{numbers, compress}{natbib}
\usepackage[final]{style/neurips_2025}

\usepackage[utf8]{inputenc} % allow utf-8 input
\usepackage[T1]{fontenc}    % use 8-bit T1 fonts
\usepackage{hyperref}       % hyperlinks
\usepackage{url}            % simple URL typesetting
\usepackage{booktabs}       % professional-quality tables
\usepackage{amsfonts}       % blackboard math symbols
\usepackage{nicefrac}       % compact symbols for 1/2, etc.
\usepackage{microtype}      % microtypography
\usepackage{xcolor}         % colors

\usepackage{microtype}
\usepackage{graphicx}
\usepackage{subfigure}
\usepackage{caption}
\usepackage{subcaption}
\usepackage{booktabs} % for professional tables
\usepackage{wrapfig}
\usepackage{xcolor}

\usepackage{amsmath}
\usepackage{amssymb}
\usepackage{mathtools}
\usepackage{amsthm}

\usepackage{algorithm}
\usepackage{algorithmic}

\usepackage{subcaption}

\usepackage[textsize=tiny]{todonotes}

\newcommand{\ourname}{NAtS}

\newcommand{\dHead}{\mathit{d}_{head}}

\newcommand{\attQ}{Q}
\newcommand{\attK}{K}
\newcommand{\attV}{V}
\newcommand{\attO}{O}
\newcommand{\attMap}{A}
\newcommand{\attMSK}{M}

\newcommand{\GlobalToken}{Global Token}
\newcommand{\LocalToken}{Local Token}
\newcommand{\SlidingWindowToken}{Sliding Window Token}

\newcommand{\GlobalTokens}{\GlobalToken s}
\newcommand{\LocalTokens}{\LocalToken s}
\newcommand{\SlidingWindowTokens}{\SlidingWindowToken s}
\newcommand{\bsz}{\mathit{B}}
\newcommand{\seqlen}{\mathit{L}}
\newcommand{\seqlenCache}{\seqlen_{\text{cache}}}
\newcommand{\dData}{\mathit{d}}
\newcommand{\nopts}{\mathit{N_{\text{opts}}}}
\newcommand{\nheads}{\mathit{H}}

\newcommand{\SlidingWindowSize}{\textit{W}}

\newcommand{\GlobalTokenSize}{$Size_G$}
\newcommand{\LocalTokenSize}{$Size_L$}

\newcommand{\AttScoreLayer}{Attention Score Layer}

\newcommand{\EndSeqHard}{G}

\newcommand{\EndSeqHardTrue}{I^{G_{\text{next}}}}

\newcommand{\SparseRegularizedPar}{\lambda}

\newcommand{\tokenstates}{\alpha}

\newcommand{\nlayers}{\mathit{N}}

\newcommand{\vQ}{\mathbf{Q}}
\newcommand{\vK}{\mathbf{K}}
\newcommand{\vV}{\mathbf{V}}
\newcommand{\vdQ}{\mathbf{dQ}}
\newcommand{\vdK}{\mathbf{dK}}
\newcommand{\vdV}{\mathbf{dV}}
\newcommand{\vS}{\mathbf{S}}
\newcommand{\vdS}{\mathbf{dS}}
\newcommand{\vP}{\mathbf{P}}
\newcommand{\vdP}{\mathbf{dP}}

\newcommand{\vO}{\mathbf{O}}
\newcommand{\vdO}{\mathbf{dO}}

\newcommand{\diag}{\mathrm{diag}}

\usepackage{amsmath,amsfonts,bm}

\def\eqref#1{equation~\ref{#1}}
\def\1{\bm{1}}

\DeclareMathAlphabet{\mathsfit}{\encodingdefault}{\sfdefault}{m}{sl}
\SetMathAlphabet{\mathsfit}{bold}{\encodingdefault}{\sfdefault}{bx}{n}

\newcommand{\softmax}{\mathrm{softmax}}

\usepackage{hyperref}
\usepackage{url}
\usepackage{amsmath}
\usepackage{caption}
\usepackage{subcaption}
\usepackage{nicematrix}
\usepackage{graphicx}
\usepackage{wrapfig}
\usepackage{paralist}
\usepackage[inline]{enumitem}
\usepackage{float}

\title{Neural Attention Search}

\author{%
  Difan Deng\\
  Leibniz University Hannover\\
   \href{mailto:d.deng@ai.uni-hannover.de}{\texttt{d.deng@ai.uni-hannover.de}}
   \And
   Marius Lindauer \\
   Leibniz University Hannover\\
   L3S Research Center\\
   \href{mailto:m.lindauer@ai.uni-hannover.de}{\texttt{m.lindauer@ai.uni-hannover.de}}
}

\begin{document}

\maketitle

\begin{abstract}
We present Neural Attention Search (\ourname), an end-to-end learnable sparse transformer that automatically evaluates the importance of each token within a sequence and determines if the corresponding token can be dropped after several steps. To this end, we design a search space that contains three token types: \begin{inparaenum}[(i)]
    \item \GlobalTokens{} will be preserved and queried by all the following tokens;
    \item \LocalTokens{} survive until the next global token appears; and
    \item \SlidingWindowTokens{} have an impact on the inference of a fixed size of the next following tokens. 
\end{inparaenum}
Similar to the One-Shot Neural Architecture Search approach, this token-type information can be learned jointly with the architecture weights via a learnable attention mask. Experiments on both training a new transformer from scratch and fine-tuning existing large language models show that \ourname{} can efficiently reduce the KV cache size and the inference costs for the models while maintaining the models' performance. 
\end{abstract}

\section{Introduction}
The ability to understand and infer from long-context information is crucial for many tasks such as long document summarization~\cite{zhang-arxiv23d} and question answering~\cite{dasigi-naacl21a, kocisky-tacl18a}, code generation~\cite{guo-icml23a,liu-iclr24b} or multi-round dialogues~\cite{yi-arxiv24a}. Thanks to the ability to query the information from any position of the historical sequence, transformer-based large language models~\cite{anthropic-24a, brown-neurips20a, gemini-arxiv24a, grattafiori-arxiv24a1,   jiang-arxiv23a} extend their context length up to millions of tokens. 

However, querying information from historical sequences requires a complexity of $\mathcal{O}(\seqlen^2)$ w.r.t. the input sequence length $\seqlen$. KV caching could reduce this time complexity to $\mathcal{O}(\seqlen)$ by storing all the historical KV values. Nevertheless, with the increasing model size of recent LLMs, even the $\mathcal{O}(\seqlen)$ time-wise and memory-wise complexity could become a bottleneck during inference time. 

Indeed, not all the tokens in a sequence are equally important~\cite{kim-kdd22a}. Many of them are redundant and do not contribute to the final output. Humans can recognize this information without pre-defined fixed rules and summarize or discard the context information into much smaller content. Transformers could also learn this ability implicitly: Many tokens in the attention map might only have very low weights~\cite{zhang-neurips23a} and only have little influence on the final predictions. However, as the transformer learns this information implicitly, we might not know how the important tokens would be distributed in the context. Selecting these tokens and recognizing the attention distributions might require extra human experts' knowledge by either looking at the attention maps~\cite{feng-arxiv24a, liu-neurips23a, zhang-icml24a, zhang-neurips23a} or applying specific fixed rules~\cite{beltagy-arxiv20a, chen-arxiv24b, child-arxiv19a, ge-iclr24a, xiao-arxiv23a}. Since this knowledge is already contained in the transformer models, we could also ask the model to evaluate the importance of each token and learn to predict the optimal type for the given input tokens automatically. 

Unlike prior works that rely on human expertise or predefined rules to identify important tokens~\cite{cai-arxiv24a, feng-arxiv24a, feng-arxiv25a, ge-iclr24a, li-arxiv24a, xiao-arxiv23a, xiao-arxiv24a, zhang-neurips23a}, we propose a novel approach to evaluate the importance of each token by assigning different roles to each of the tokens. For example, some tokens will be preserved until the end, while other tokens might only survive for a short amount of time. These roles measure the importance of each token and determine if it would survive within the next few tokens. Rather than pre-defining a set of fixed rules for each token, we ask the model to learn this information automatically. Finding the optimal role for each token is similar to the process of neural architecture search~\cite{elsken-automlbook19a, white-arxiv23a}, where an optimal architecture is searched by the optimizer for a given task. Thereby, in this work, we jointly optimize the choice of each token and the model weights by constructing a learnable attention mask. Our approach is implicitly motivated by the one-shot neural architecture search approach~\cite{dong-cvpr19a, liu-iclr19a, pham-icml18a}, where the model parameters and architecture parameters are jointly optimized during the search process. However, our approach, Neural Attention Search (\ourname), searches for the optimal token roles jointly with the attention weights.

Our contributions are as follows:
\begin{compactenum}
\item We propose Neural Attention Search (\ourname), an end-to-end learnable sparse attention framework that automatically searches for the optimal roles for each token in the input sequence. 
\item We introduce different token roles in our search space that can be later combined to construct a learnable attention mask and then jointly optimized with the model weights in \ourname. 
\item We show that \ourname~ could efficiently reduce the KV cache required during inference time while maintaining most of the models' performance. 
\end{compactenum}

By automatically learning to focus on the most relevant information, \ourname{} paves the way for more efficient and scalable inference with LLMs in long-context applications.

 \section{Background and Related Work}
\subsection{Attention maps for transformers~\label{sec:attmap}}
Transformers~\citep{vaswani-neurips17a} computes the correlation between different tokens by mapping the input sequences into three variables, $\attQ$, $\attK$, and $\attV$. An attention map $\attMap$ is first computed by pairing $\attQ$ and $\attK$ and then scaled and normalized with a softmax function. Finally, this attention map $\attMap$ is multiplied by~$\attV$. Additionally, a mask $\attMSK$ is attached to the Attention Map to guide the attention to focus only on certain tokens. This can be expressed as 
\begin{equation}
     \attMap  = \frac{\attQ\attK^T}{\sqrt{\dHead}}, \hspace{10mm}
    \attO  = \softmax(\attMap  + \attMSK^{add})\attV\label{eq:attmap}
\end{equation}

The additive attention mask $\attMSK^{add}_{i,j} \in \{ -\inf, 0 \}$ controls if the transformer needs to construct the correlation between $\attQ_i$ and $\attK_{j}$.  A widely used mask $\attMSK^{add}$ used in transformers is a causal mask~\citep{vaswani-neurips17a}, where the upper parts of the mask are filled with $-\inf$; each token can only query the token information before it. 

Computing the attention map of a sequence with length $\seqlen$ requires a complexity of $\mathcal{O}(\seqlen^2)$ for both time and memory. This complexity can be lowered to $\mathcal{O}(\seqlenCache)$ during inference time, with  $\seqlenCache$ being the length of the KV cache. However, as an expense, the cost of storing the KV cache gradually becomes unnegligible with the increasing size of large language models and the context length they could handle. Therefore, lots of work has been proposed to reduce the storage and computation overheads during the inference time~\cite{li-arxiv24b}.

To reduce the computational costs of an LLM on long context information, LLMLingua~\cite{pan-arxiv24a} trains another smaller model to compress the input texts. Many other studies work on reducing the KV cache size by deciding which tokens to evict or preserve with the past attention maps. Recent works such as H2O~\citep{zhang-neurips23a}, Scissorhand~\cite{liu-neurips23a}, FastGen~\cite{ge-iclr24a}, KeyFormer~\cite{adnan-arxiv24a}, SnapKV~\cite{li-arxiv24a}, PyramidKV~\cite{cai-arxiv24a}, AadaKV~\cite{feng-arxiv24a}, ChunkKV~\cite{liu-arxiv25a}, and CriticalKV~\cite{feng-arxiv25a} all apply different rules to identify the importance of each token and remove all the remaining tokens. Additionally, instead of simply removing the unimportant tokens, we can also merge them into the existing tokens~\cite{wan-arxiv24a, zhang-icml24a}. 

%CAM~\citep{zhang-icml24a} achieves this by merging the to-be-evicted tokens into the remaining tokens. 

Other works do not aim at reducing the KV cache size, but only select a subset of KV caches to compute the attention outputs, including Quest~\cite{tang-icml24a}, TokenButler~\cite{akhauri-arxiv25a}, AttentionPredictor~\cite{yang-arxiv25b}, and XAttention~\cite{xu-arxiv25a}. However, these works still need to maintain the entire KV cache in the GPUs. 
This still results in a huge GPU memory usage overhead. To alleviate this issue, some other works like InfLLM~\cite{xiao-arxiv24b}, HIP~\cite{lee-arxiv25a}, and ShodwKV~\cite{sun-arxiv24b} offload the KV cache to CPUs and only load the KV caches that are relevant for the current token predictions to GPUs. 

While these existing methods offer various ways to reduce KV cache size, they often rely on inflexible predefined rules and potentially inaccurate heuristics based on past attention maps. \ourname, in contrast, introduces a novel approach that learns to dynamically assign token roles during inference, enabling a more adaptive and efficient use of the KV cache. By treating token role assignment as an optimization problem, \ourname{} leverages principles from neural architecture search to jointly optimize token roles and model weights, leading to a more flexible and powerful attention mechanism.

%Alternatively, past attention maps can decide which tokens to evict or preserve. H2O~\citep{zhang-neurips23a}  and Scissorhand~\cite{liu-neurips23a} both identify the importance of each token based on their contributions to the attention maps and only preserve the most important tokens. Other approaches such as FastGen~\cite{ge-iclr24a}, SnapKV~\cite{li-arxiv24a}, PyramidKV~\cite{cai-arxiv24a}, and AadaKV~\cite{feng-arxiv24a} all applied pre-defined fixed rules to identify the important KV values. Additionally, instead of simply removing the unimportant tokens, we can also merge them into the existing tokens. CAM~\citep{zhang-icml24a} achieves this by merging the to-be-evicted tokens into the remaining tokens. 

\subsection{Sparse Attention}
Sparse attention works on computing only a fraction of the attention map to reduce the computation costs and KV cache sizes with either pre-defined fixed patterns, including Sparse Transformers~\cite{child-arxiv19a}, Longformer~\cite{beltagy-arxiv20a}, StreamingLLM~\cite{xiao-arxiv23a}, and DuoAttention~\cite{xiao-arxiv24a}. Alternatively, this can be adjusted by special embedding tokens such as 
Longcoder~\cite{guo-icml23a} and SepLLM~\cite{chen-arxiv24b}. Additionally, one can also determine the types of sparse attention with a set of proxy values, including MInference~\cite{jiang-neurips24a}, FlexPrefill~\cite{lai-iclr25a}, SeerAttention~\cite{gao-arxiv24a}, SpargeAttention~\cite{zhang-arxiv25a}, and XAttention~\cite{xu-arxiv25a}. However, these approaches aim to approximate the attention output of each layer from a pre-trained dense transformer that only involves the QK matrices. In contrast, \ourname{} learns the token roles directly from the final loss function and does not necessarily need to approximate the existing attention output. Hence, \ourname{} can be applied to both training from scratch or approximating the outputs from an existing model. Additionally, the existing approaches can only be applied to approximating the attention maps during the pre-filling stages and still need to preserve the entire KV cache during the decoding stage, making them less profitable for reasoning models where the model needs to generate lots of tokens during the decoding stage~\cite{deepseek-arxiv25a}. In contrast, \ourname{} only checks the roles for each token and does not necessarily need to link to the actual attention map values. Hence, it can be applied to both pre-filling and decoding stages while keeping the KV cache sizes to a low level.

In contrast to the fine-tuning-free approaches above, we can also fine-tune the target model to achieve the desired sparsity. This includes  Adaptively Sparse Attention~\cite{anagnostidis-neurips23a} and  Dynamic Memory Compression~\citep{nawrot-icml24a}.  However, both approaches rely on expensive cumulative productions during training time, making their approach less applicable to long-context scenarios. LLMLingua2~\cite{pan-arxiv24a} trains another model to compress the target texts.  Additionally, ~\citet{kim-kdd22a} asks the model to learn to drop the tokens with lower attention scores. Landmark Attention~\cite{mohtashami-neurips23a} inserts landmark tokens after certain time steps and applies these landmark tokens to summarize the information of the tokens before that landmark token and recover. SPARSEK~\citep{lou-arxiv24a} only selects a constant number of KV pairs for each query by introducing a differentiable SparseK operator. Mixture-of-Depth~\cite{rapose-arxiv24a} only passes a limited number of tokens to each transformer layer. Native Sparse Attention~\cite{yuan-arxiv25a} and MoBA~\cite{lu-arxiv25a} all select a subset of the KV values to compute the attention masks and therefore reduce the computational overhead. However, these approaches still require preserving the entire KV cache values and only reducing computational costs during decoding stages. 

%Adaptive sparse attention ~\cite{anagnostidis-neurips23a}  shows that this sparsity can also be learned through sparse sigmoid functions.  Dynamic Memory Compression~\citep{nawrot-icml24a} tries to merge multiple tokens into one single token dynamically. However, these approaches rely on expensive cumulative productions during the training time, especially in long-context scenarios. Additionally, ~\citet{kim-kdd22a} asks the model to learn to drop the tokens with lower attention scores. Landmark Attention~\cite{mohtashami-neurips23a} inserts landmark tokens after certain time steps and applies these landmark tokens to summarize the information of the tokens before that landmark token and recover. SPARSEK~\citep{lou-arxiv24a} only selects a constant number of KV Paris for each query by introducing a differentiable SparseK operator. Mixture-of-Depth~\cite{rapose-arxiv24a} only passes a limited number of tokens to each transformer layer. Native Sparse Attention~\cite{yuan-arxiv25a} and MoBA~\cite{lu-arxiv25a} all select a subset of the KV values to compute the attention masks and therefore reduce the computational overhead. However, these approaches still require preserving the entire KV cache values and only reducing computational costs during decoding stages. 

MoA~\cite{fu-arxiv24a} proposes to search for the optimal sparse attention for different heads, which share similar ideas to ~\ourname{}. However, MoA's search space only contains streaming LLM.%, and it only searches for the sizes of the initial tokens and sliding window tokens.
This head-level search space of MoA is much smaller than our token-level search space, which might restrict the model's expressibility under even smaller budgets.

%Landmark Attention~\cite{mohtashami-neurips23a} inserts landmark tokens after certain time steps and applies these landmark tokens to summarize the information of the tokens before that landmark token and recover. 

\subsection{Neural Architecture Search}
Designing a neural architecture for a specific task might require a lot of trial and error. Neural Architecture Search (NAS) automates this process by searching in a pre-defined search space~\cite{elsken-jmlr19a}. Previous work on NAS mainly focused on searching within a discrete search space by sampling a new architecture from the search space, evaluating its performance, and updating the optimizers~\cite{zoph-iclr17a, zoph-cvpr18a}. %However, training every network from scratch requires lots of GPU hours. 
One-Shot NAS~\cite{pham-icml18a} approaches instead share the same weights of the operators with respect to all the architectures in the search space. This allows to jointly optimize the architectures and weights. DARTS~\cite{liu-iclr19a} and GDAS~\cite{dong-cvpr19a} further relax the discrete search space into continuous values to optimize the architecture parameters with gradient descent. The One-Shot NAS approach allows the optimizers to efficiently search for the optimal architecture within a huge search space. Similarly, \ourname~ has multiple options for each token as the search space and is able to search for the optimal token types jointly with the model weights. However, unlike One-Shot NAS approaches that consider the optimization problem as a bilevel optimization problem and optimize the model weights and architecture weights alternatively, we optimize the token state information and model weights within one forward and backward pass. This is similar to a mixture-of-experts (MOE) model~\cite{fedus-jmlr23a, shazzer-iclr17a}. However, instead of distributing data across all experts uniformly, we only select one expert for each token and assign all data to that expert.

\section{The \ourname{} Approach~\label{sec:approach}}
In this section, we first introduce all the candidate token types in our search space. We then show that we can construct a learnable attention mask with the choice of each token type. Finally, we can efficiently reduce the KV cache size by dropping the unnecessary tokens during inference.

\subsection{Search Space Design~\label{sec:searchspace}}
Not all the tokens in a sequence are equally important. %Some of the tokens might contain important information, and they should be preserved that will be further queried by the following tokens. Some tokens might only contribute to the prediction for the next few tokens. 
Just like a paragraph is composed of multiple sentences, a sequence can be divided into multiple sub-sequences containing different information; some tokens might only be required within these sub-sequences. Hence, we design a search space for each token's role within the sequence and ask the model to automatically learn the optimal role for each token in the sequence. 

We first define \textbf{\GlobalToken{}s} as tokens containing important information that need to be preserved for the following predictions. \citet{liu-neurips23a} and \citet{zhang-neurips23a} showed that only a small fraction of the tokens contribute to most of the attention scores for self-attention computation. These tokens need to be preserved for models to recall the global information that helps to predict the next token. In vanilla transformer models, all the tokens are \GlobalTokens. 

\GlobalTokens{} will not be evicted during inference time. Therefore, we should maintain as few \GlobalTokens{} as possible to ensure inference efficiency. Each \GlobalToken{} should not only preserve the information within that position. Ideally, it should also be able to summarize the information from previous sequences~\cite{ chen-arxiv24b, guo-icml23a}. Hence, we split the entire sequence with the \GlobalTokens{} into multiple sub-sequences, with each sub-sequence ending with one \GlobalToken. Each \GlobalToken{} only needs to summarize the information from its sub-sequences and the previous \GlobalTokens{}. 

\textbf{\LocalToken{}s} only survive until the next \GlobalToken{} appears. Therefore, models will have the full attention within each sub-sequence to summarize the local sub-sequence information into the \GlobalToken{} which is located at the end of the sub-sequence. Meanwhile, the model will be sparse within the input sequence.  \LocalToken{}s are considered as tokens that only provide lower-level information that helps the model to understand the local context information, but might not help further outside this context. This provides the model with a flexible interface to control its sparsity based on the input context information.

Only the \GlobalTokens{} and \LocalTokens{} might control the sparsity at a low granularity level. E.g., assuming that one input sequence is highly localized, each token only has a high correlation with itself or a few neighboring tokens. In this case, they are all similar and are assigned with the same token type. However, none of the \GlobalToken{} and \LocalToken{} could sparsify this attention map efficiently: if all the tokens are classified as \LocalTokens, the input sequence will only be considered as one single subsequence, and all the \LocalTokens{} will be equivalent to the \GlobalTokens. 

Hence, we introduce \textbf{\SlidingWindowToken}. 
\SlidingWindowTokens{} will only be preserved for the next \SlidingWindowSize{} time steps and were previously considered as one of the most popular sparse attention approaches~\cite{child-arxiv19a, ge-iclr24a, xiao-arxiv23a, zhang-neurips23a}.

\begin{figure*}[h]
\centering
\subfigure[]{
     \centering
     \includegraphics[width=0.17\textwidth]{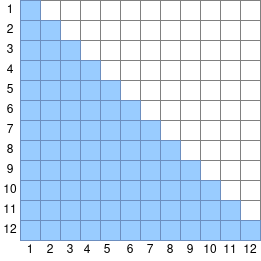}
         \label{fig:FullAtt}
    }
    \hfill
\subfigure[]{
     \centering
     \includegraphics[width=0.17\textwidth]{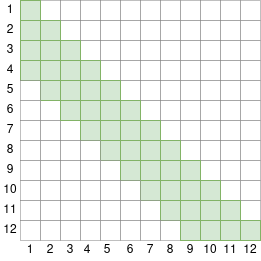}
    \label{fig:LocalAtt}}
    \hfill
\subfigure[]{
     \centering
     \includegraphics[width=0.17\textwidth]{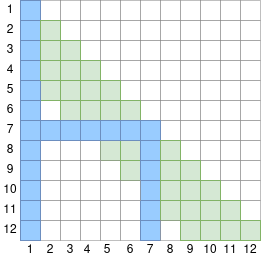}
    \label{fig:LongFormer}}
    \hfill
\subfigure[]{
     \centering
     \includegraphics[width=0.17\textwidth]{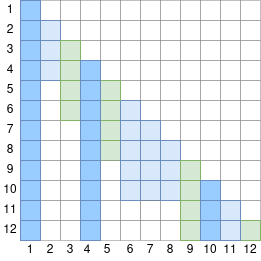}
    \label{fig:LearnedAttention}}
    \begin{subfigure}
    \centering
     \includegraphics[width=0.1\textwidth]{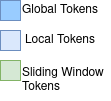}
     \end{subfigure}
    \caption{A comparison between different casual attention maps. \ref{fig:FullAtt}: The full attention map, where each token is connected to the tokens before it. \ref{fig:LocalAtt}: the local attention with sliding windows 3, every token will only get access to the information of the 3 tokens ahead. \ref{fig:LongFormer} Longformer, besides the local attention, the first, 6th and 9th tokens are the pre-defined global tokens. \ref{fig:LearnedAttention}: \ourname ~dynamically 
    learns the optimal role for each token and constructs a learnable mask based on the tokens' roles. 
    \label{fig:attentinos}}
\end{figure*}

In contrast to other causal attention maps, Figure~\ref{fig:LearnedAttention} illustrates an exemplary attention mask constructed by the choices of different token types. In this case, we define the sliding window size as 4. Token 1, 4, 10 act as \GlobalTokens; Tokens 2, 6, 7, 8, 11 are \LocalTokens; Token 3, 5, 9, 11 are \SlidingWindowTokens{}. The \GlobalTokens{} splits the entire sequence into three subsequences that end at 4, 10, and the last index, respectively. Hence, Token 2 will only be required by Token 3 and~4. This rule applies the same for Token 6, 7, and 8, where they only interact with the tokens within the same subsequence. For the sliding window tokens, only the next three tokens could query their information, regardless of whether these tokens belong to the same sub-sequence. Only 6 out of 12 tokens are involved during inference time to make the next token prediction.

Combining different token types could already cover most of the sparse attention variables. For instance, assigning \GlobalTokens{} to the first few tokens and \SlidingWindowTokens{} to all the other tokens results in streaming LLMs~\cite{xiao-arxiv23a} and MoA~\cite{fu-arxiv24a}. If we further set all the tokens in some heads as \GlobalTokens{}, we will get DuoAttention~\cite{xiao-arxiv24b}. Assigning different tokens as \GlobalTokens{} results in the Vertical patterns~\cite{jiang-neurips24a} that appear in many Token eviction strategies~\cite{li-arxiv24a, liu-neurips23a}. Since different layers or heads might have different numbers of global tokens, patterns like pyramidkv~\cite{cai-arxiv24a} and adakv~\cite{feng-arxiv24a} can be easily derived from our search space. For a transformer network with $\nlayers$ layers and $\nheads$ attention heads that needs to process one sequence of length $\seqlen$, given the three token types defined above, there are $3^{\seqlen \times \nheads \times \nlayers}$ possible configurations in our search space. It is prohibitive to check all the combinations in such a huge search space, with the increasing sequence length that a model receives. %Following the previous work on one-shot neural architecture search~\cite{dong-cvpr19a, liu-iclr19a}, we propose an end-to-end approach that allows us to search for the optimal token type combinations and the network weights jointly. 
Here, we propose an end-to-end approach to search for the optimal token types together with the network weights jointly.

\subsection{Searching for the Optimal Token Types~\label{sec:natstrain}}
Searching for the optimal token types within a sequence is similar to searching for the optimal architectures for a given task in neural architecture search~\cite{elsken-jmlr19a, white-arxiv23a}. Following GDAS~\cite{dong-cvpr19a}, we apply the Gumbel-Softmax trick~\cite{jang-iclr17a, maddison-iclr17a} to sample from the discrete search space. The Gumbel-Softmax trick allows the gradient to be backpropagated through the discrete choice of the token types. 

Specifically, we first use a linear layer (which we call \AttScoreLayer) that maps each the input tensor for an attention layer $X \in \mathbb{R}^{\dData}$ to the likelihood for each option: $\tokenstates \in \mathbb{R}^{(\nheads * \nopts)} = \text{Linear}(\mathbf{x})$, where $\nheads$ is the number of KV heads and $\nopts$ is the number of options in the search space. The type $\tokenstates$ for each token is then sampled from a Gumble-Softmax function based on the likelihood values. Given that the number of options $\nopts$ is normally much smaller than the model head dimensions, this linear layer only introduces minimal overhead compared to the QKV linear layers.

We now construct a learnable attention mask $\attMSK$ with a series of sampled token states. However, the additive mask in Eq.~\ref{eq:attmap} will take $-\inf$ values,  resulting in invalid gradient information. Hence, we use the token information to construct a multiplicative attention mask $\attMSK^{\text{mul}} \in \{0, 1\}~\cite{rao-neurips21a}$\footnote{For the sake of brevity, we will use $\attMSK$ instead of $\attMSK^{\text{mul}}$ in the following part of this paper.}:
\begin{equation}
    \attO = \frac{e^{A} \odot \attMSK^{\text{mul}}}{\sum_j e^{A_{.,j}} \odot \attMSK^{\text{mul}}_{., j}}\attV\label{eq:attoutmul}
\end{equation}

The attention mask columns for \GlobalTokens{} and \SlidingWindowTokens{} can be directly constructed since they will survive for a fixed number of steps. However, the mask for \LocalTokens{} $\attMSK^{L}_{i,j}$ is controlled by both the distribution from \LocalTokens{} and \GlobalTokens{} as \LocalTokens{} will survive until the next \GlobalToken{} appears. In other words, to make $\attMSK^{L}_{i,j} (j>i)$ a valid value, no \GlobalToken{} should appear between $i$ and $j$. 

Formally, the attention masks can be created as follows:
\begin{align}
    \attMSK^{G}_{i, j} & = 1\label{eq:mskglobal} \\
    \attMSK^{SW}_{i,j} & = 
    \begin{cases} 
    1 & \text{if}\ j \leq i + \SlidingWindowSize \\
    0 & \text{if}\ j > i + \SlidingWindowSize
    \end{cases}\label{eq:msksw}\\
    \attMSK^{L}_{i, j} & = \prod_{n=j+1}^{i-1}(1-\EndSeqHard_{n}) \label{eq:msklocal}
\end{align}
where \SlidingWindowSize{} is the sliding window size and $\EndSeqHard_{n}$ a \GlobalToken{} at Position $n$. We then construct the attention masks based on the type of each token. After that, we mask out the upper triangular part of the mask to ensure its causality.

In practice, we first collect the index of the next global token  
$\EndSeqHardTrue_i := \min (\{k | k >= i \land \EndSeqHard_k = 1 \})$ and rewrite Eq.~\ref{eq:msklocal} as:
\begin{align}
    \attMSK^{L}_{i, j} = \begin{cases}
        1 & \text{if}\  j \leq \EndSeqHardTrue_i \\
        0 & \text{if}\  j > \EndSeqHardTrue_i
    \end{cases} \label{eq:msklocalfwd}
\end{align}

During the backward process, we collect the column-wise gradient values from our attention masks and apply these gradients to update the parameters from our \AttScoreLayer{}. Different from other approaches that approximate the attention maps or attention outputs from an existing  model~\cite{akhauri-arxiv25a, fu-arxiv24a, yang-arxiv25b}, ~\ourname{} can be learned directly from the target labels, and thus, can be applied for both pre-training a new sparse transformer from scratch or fine-tuning an existing transformer. To control the attention map's sparsity, we introduce a small regularization value $\SparseRegularizedPar$ that is directly applied to the gradient for \GlobalToken{} and \LocalTokens{} to encourage more \SlidingWindowTokens. This regularization value $\SparseRegularizedPar$ can be considered as a value similar to a soft threshold, pushing the tokens whose contributions in the attention maps outside the scope of the sliding window sizes with lower levels towards \SlidingWindowTokens{}. Hence, we could efficiently reduce the corresponding overall computational costs.

We show how the backward gradients are computed and how the $\SparseRegularizedPar$ values control the attention map sparsity in the appendix ~\ref{apd:backward}.

These rules are then integrated in FlashAttention~\cite{ dao-iclr24a, dao-neurips22a} to avoid explicitly computing the attention masks during the forward pass. In addition to the transformer computation, we only need to collect the next \GlobalToken{} indices $\mathbf{\EndSeqHardTrue}$ (with a complexity of $\mathcal{O}(N)$) and then mask out the attention map values with the masks defined above. At the same time, we also skip all the computation blocks that do not contain any valid values (i.e., all the items in the mask of that block are 0) during the online attention process. Further details can be found in the appendix \ref{sec:nats-fa2}

\subsection{Efficient Inference with Different Token Types~\label{sec:cacheupdate}}
\ourname{} can be applied to both pre-filling and decoding stages. During the pre-filling stage, we first compute the attention values for all the global and local tokens. We then compute attention map values for the sliding window tokens separately to fully utilize the parallelism of GPU architectures.

During the decoding stage, we dynamically map the input feature maps to the corresponding token types and discard the tokens no longer required by the following tokens once a new token arrives.
The \SlidingWindowTokens{} only survive for a fixed number of time steps. We preserve a queue in the KV cache space that only stores the most recent $\SlidingWindowSize$ tokens and mask out the non-\SlidingWindowTokens: when new tokens come, we place them after the tail of the queue to replace the tokens older than $\SlidingWindowSize$. 

Similar to the vanilla transformer forward process, when new tokens arrive, we concatenate them with the existing KV caches, generating new masks and computing the attention output. After that, we start our post-update process: we first check the state of each token to decide if we can remove them or keep them in the cached values. Since different heads might disagree on the types of the same token, we record both the sizes for \GlobalTokens~(\GlobalTokenSize) and \LocalTokens~(\LocalTokenSize) for all the heads. New \SlidingWindowTokens{} do not change these sizes since they will always be placed separately. However, when a new \GlobalToken{} for any head arrives, we remove all the \LocalTokens{} from the corresponding heads and place the new \GlobalToken{} right after the existing \GlobalTokens{} and then update our \GlobalTokenSize{} and \LocalTokenSize{} accordingly. The same strategy is applied when new $\LocalTokens$ arrive: we place them at the end of the \LocalTokens{} and increase the number for \LocalTokens. A detailed updating process can be found under Appendix~\ref{sec:inferencedetails}.

\begin{wrapfigure}[11]{r}{0.45\textwidth}
 \vspace{-2.5mm}
    \centering
    \includegraphics[width=1.0\linewidth]{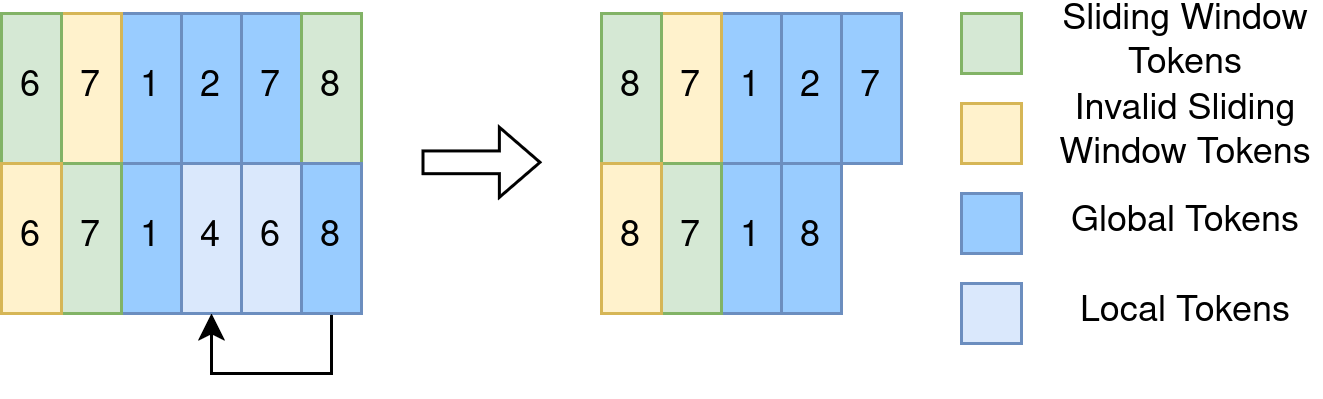}
    \caption{An example of how caches are updated within \ourname{} when new tokens arrive with a model containing two heads. The two rows represent different heads.}
    \label{fig:cacheUpdate}
\end{wrapfigure}

Figure~\ref{fig:cacheUpdate} illustrates an exemplary case to update the KV caches in \ourname{} with a sliding window size of 3.
The first two places store the most recent tokens and use a mask to mask out the non-\SlidingWindowTokens~(yellow tokens). Since Token~8 for Head 1 is \SlidingWindowToken{} and \GlobalToken{} for Head 2. We first move both tokens to the beginning of our cache to replace the old one. After that, we drop Token~8 in Head 1 since it has already moved to the sliding window caches. Then, since Token 8 in Head~2 is a \GlobalToken, we drop all local tokens after the last \GlobalToken~(Token 1). Hence, 
we place Token 8 after Token~1 and remove Tokens 4 and 6. The \GlobalTokenSize{} is then updated from $[5,3]$ to $[5,4]$ and the \LocalTokenSize{} is updated from $[0, 2]$ to $[0,0]$: both new tokens are merged into the existing tokens, we do not need the extra space to store them. 
% we do not need the extra space to store these new tokens.  %Since both new tokens are merged into the existing tokens, we do not need the extra space to store these new tokens. 

\section{Experiments~\label{sec:exp}}
We implement \ourname{} based on the Flash Attention 2~\cite{dao-iclr24a} implementation on \href{https://triton-lang.org/main/getting-started/tutorials/06-fused-attention.html}{triton}. %All the operations that we proposed in Section~\ref{sec:natstrain} have at most $\mathcal{O}(\seqlen)$ complexity.  
In our experiments, we first train \ourname{} parameters jointly within a transformer model from scratch. We then apply \ourname{} to fine-tune a large language model to show that \ourname{} could efficiently reduce the KV cache size required while maintaining the model performance. The codes for ~\ourname{} implementation and experimental designs can be found under: \url{https://github.com/automl/NeuralAttentionSearch}

\subsection{Training a Transformer From Scratch~\label{sec:trainfromscratch}}
%We first apply \ourname{} to train a GPT2 small style~\cite{radford-arxiv21a} transformer model from scratch. Following the setting from NanoGPT~\cite{karpathy-github22a}, this model has 128M parameters with 12 layers and 12 heads with a hidden dimension of 768. Instead of the learnable position encoding, we apply rotary embeddings~\cite{su-nc24a} to each transformer layer. We train this model on the PG-19 Language Modeling Benchmark~\cite{rae-iclr20a}. This dataset contains books extracted from Project Gutenberg~\cite{gutenberg} with about 2B tokens in the training sets. %We train all models with a context length of 1024 and a batch size of 480 (using gradient accumulation). We train them for $600\,000$ iterations on a computation node with four Nvidia H100-PCIe GPUs and evaluate them on the test sets of PG19 with a context length of 1024. %Further details on the hyperparameters can be found in the appendix. 

We first apply \ourname{} to train a GPT2 small style~\cite{radford-arxiv21a} transformer model with 128M parameters from scratch on the PG-19 Language Modeling Benchmark~\cite{rae-iclr20a} with four Nvidia H100-PCIe GPUs and evaluate it on the test sets of PG19 with a context length of 1024. Further details can be found under the appendix ~\ref{sec:pg19-train}. 

%Following the setting from NanoGPT~\cite{karpathy-github22a}, this model has 128M parameters with 12 layers and 12 heads with a hidden dimension of 768. Instead of the learnable position encoding, we apply rotary embeddings~\cite{su-nc24a} to each transformer layer. We train this model on the PG-19 Language Modeling Benchmark~\cite{rae-iclr20a}. This dataset contains books extracted from Project Gutenberg~\cite{gutenberg} with about 2B tokens in the training sets. We train all models with a context length of 1024 and a batch size of 480 (using gradient accumulation). We train them for $600\,000$ iterations on a computation node with four Nvidia H100-PCIe GPUs and evaluate them on the test sets of PG19 with a context length of 1024. %Further details on the hyperparameters can be found in the appendix. 

As a baseline, we train another dense transformer model under the same hyperparameter setting. During inference time, we compare \ourname{} with the following baselines besides the full Transformer: 
\begin{inparaenum}[(i)]
    \item Streaming LLM~\cite{xiao-arxiv23a} only preserves the first few starting and the most recent few tokens for future prediction. 
    \item H2O~\cite{zhang-neurips23a} first computes the attention map and only preserves the tokens with the top-k attention scores. 
\end{inparaenum}
H2O and Streaming LLM are training-free approaches that control the sparsity with pre-defined hyperparameters during inference time. %H2O needs to define the recent sliding window size and the number of Heavy Hitter (HH) tokens; Streaming LLM requires the number of tokens at the beginning of the sequence (Sink Tokens) and sliding window size. 
In contrast, \ourname{} controls this sparsity by the sparsity hyperparameter value $\SparseRegularizedPar$. Hence, we train multiple models with different $\SparseRegularizedPar$. However, in our experiments, we observe that the attention map sparsity values converge much faster than the model loss. We could quickly estimate the attention map sparsity within the first 10,000 iterations to check if this sparsity value satisfies the required sparsity and early-stop the runs that do not satisfy the requirements~\cite{jamieson-aistats16a}. 

\begin{wrapfigure}[12]{r}{0.5\textwidth}
    \centering
    \vspace{-6mm}
    \includegraphics[width=0.95\linewidth]{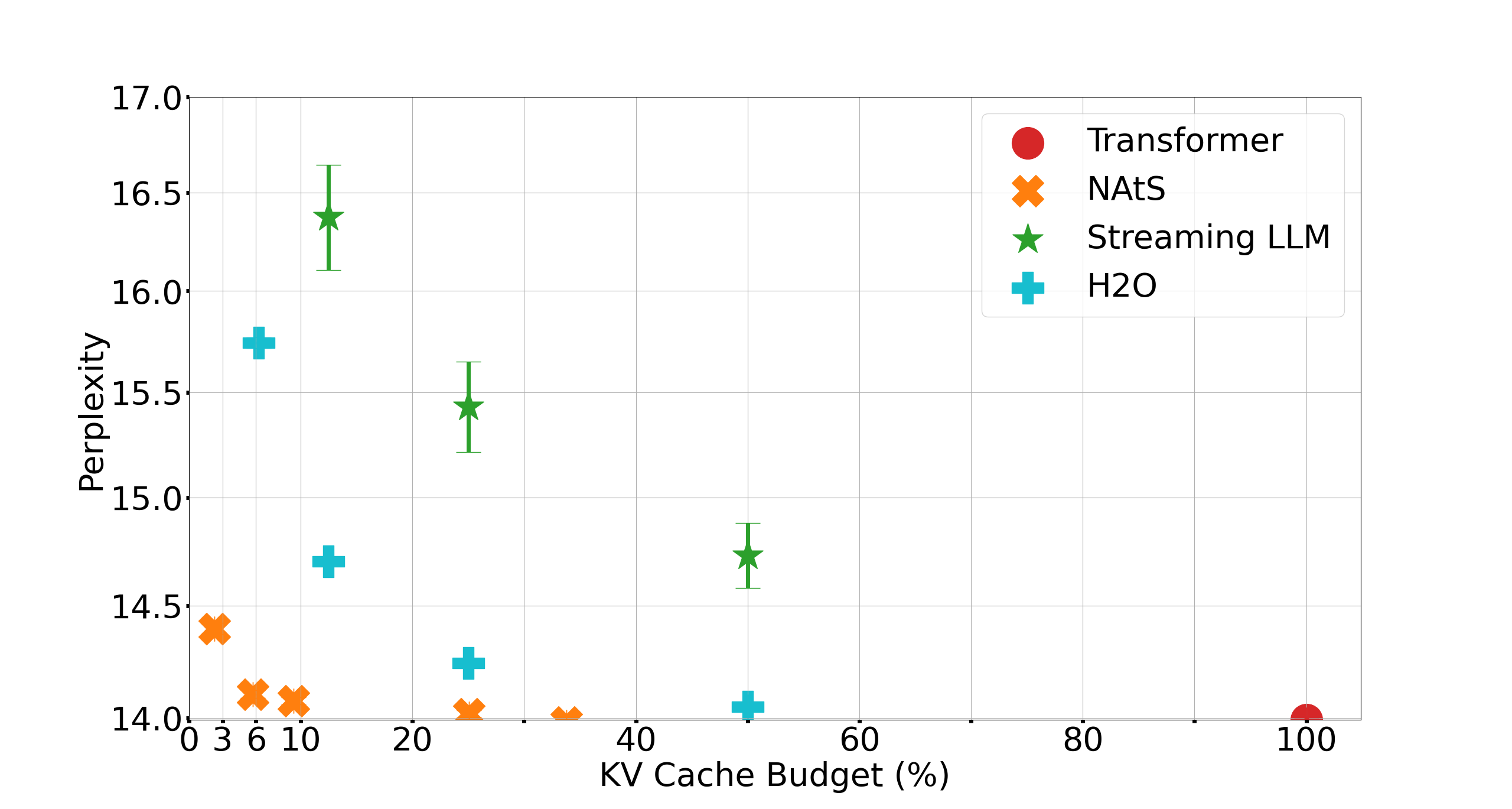}
    \caption{Perplexity vs KV Cache size under different sparsity settings $\SparseRegularizedPar$ on the PG19 dataset.}
    \label{fig:pg19perKV}
\end{wrapfigure}

 We provide different hyperparameters for H2O and Streaming LLM for different sparsity (with the sliding window size of 32, 64, 128, 256, plus the same number of HH for H2O and 64, 128, 256, 512, plus 8 Sink tokens for Streaming LLM). This results in a corresponding KV budget fraction ranging from  6.25\% to 50\%. Meanwhile, we train \ourname{} with the following $\SparseRegularizedPar$: $0, 1e-9, 5e-9, 1e-8, 1e-7$. We run each experiment with three different seeds.

Figure~\ref{fig:pg19perKV} shows the perplexity and the corresponding standard deviation (as error bars) of the PG19 test sets from different model settings across different seeds. The x-axis indicates the fraction of KV caches applied to generate the last token in the input sequence. As expected, a larger $\SparseRegularizedPar$ results in a smaller KV cache fraction, while the performance is relatively stable across different seeds. Interestingly, even if $\SparseRegularizedPar$ is set as $0$, the KV cache budget can still be reduced to around 35\% with slightly lower perplexity compared to the full attention. This might indicate that the model also tends to sparsify the input information to focus more on the relevant parts from the input data~\cite{ye-arxiv24a}. 

H2O maintains nearly lossless perplexity with a budget of 50\%. However, as the available budget decreases, ~\ourname's perplexity remains nearly the same until a KV cache size of around 10\%. In contrast, H2O and Streaming LLM start to degenerate their performance starting from a $25\%$ budget allocation. The most sparse model in \ourname{} family only contains roughly $3\%$ of the KV cache size (meaning roughly 30 cached tokens per layer on average) and achieves a lower perplexity compared to H2O with $12.5\%$ of the KV caches.

%Under the budget of around 50\%, all the approaches perform similarly to the full transformer while \ourname{} has a relatively lower perplexity and requires less budget. However, as the available budget decreases, \ourname's perplexity remains nearly the same until a KV cache size of around $10\%$. In contrast, H2O and Streaming LLM start to degenerate their performance starting from a $25\%$ budget allocation. The most sparse model in \ourname{} family only contains roughly $3\%$ of the KV cache size (meaning roughly 30 cached tokens per layer on average) and achieves a lower perplexity compared to H2O with $12.5\%$ of the KV caches and Streaming LLM with $25\%$ of the KV caches.

\subsection{Fine-Tune a Large Language Model}
We now apply \ourname{} to fine-tune an existing large language model~\cite{grattafiori-arxiv24a1, jiang-arxiv23a} in the long context scenario. We construct a new training dataset that follows the construction rules of LongBench as a training set to fine-tune LLM models with \ourname{}. Some of LongBench's tasks are collected from the test sets of the previous benchmarks. Hence, we first collect the training sets from these benchmarks and construct these datasets following the data structure in LongBench. Additionally, we also add the passkey-retrieval dataset introduced in DuoAttention~\cite{xiao-arxiv24b}. Overall, this dataset contains $7\,000$ instances with a maximal context length of $16\,000$. Further details can be found in Appendix~\ref{sec:datacollect}. 
%Overall, the longbench dataset contains $6\,436$ pieces of data with a maximum context length of $16\,000$. Further details on the dataset collection can be found in the appendix. 

% \begin{figure}
% \centering
% \subfigure{\includegraphics[width=0.235\textwidth]{figures/Exp/NIAH/niah_fullattn.png}}
% \subfigure{\includegraphics[width=0.235\textwidth]{figures/Exp/NIAH/niah_duo.png}}
% \vfill
% \subfigure{\includegraphics[width=0.235\textwidth]{figures/Exp/NIAH/niah_h2o.png}}
% \subfigure{\includegraphics[width=0.235\textwidth]{figures/Exp/NIAH/niah_streaming.png}}
% \vfill
% \subfigure{\includegraphics[width=0.235\textwidth]{figures/Exp/NIAH/niah_nats.png}}
% \subfigure{\includegraphics[width=0.235\textwidth]{figures/Exp/NIAH/niah_budget_nats1e7.png}}
% \caption{Results on Needle in a Haystack test with Llama-3-8B-Instruct-Gradient-1048k model. The context length ranges from $1\,000$ to $32\,000$ and the document depth ranges from $0$ to $1$. We assign a $25\%$ budget to each of the approaches. The first two rows are the scores (higher is better) of the full attention and the corresponding baselines. The bottom plot on the left shows the scores from \ourname{} with an average compression rate of $15\%$. The plot on the right shows the compression rate of \ourname{} across different document lengths~\label{fig:expniah}.}
% \end{figure}

We only fine-tune the \AttScoreLayer{} while keeping all the other parameters in the network fixed to approximate the original output from the corresponding base LLM. This approach is similar to DuoAttention~\cite{xiao-arxiv24a}. However, since we want the model to capture the overall context information, we update \AttScoreLayer{} with all the output from the full attention layer:
%The authors of DuoAttention first embed multiple randomly generated passkey sequences into random positions of a long context and then asked the model to recall the passkey sequence. However, DuoAttention only needs to recognize the retrieval heads for a sequence level. Our \AttScoreLayer{} has to identify the roles for each token. This requires the model to capture both long-term global information and short-term local correlation. Hence, we ask the model to make the next token prediction for all the text it has received. Then similar to ~\citet{xiao-arxiv24a}, we then train the \AttScoreLayer{s} to approximate the output from the full transformer with L1 Loss:
\begin{equation}
    \mathcal{L}_{distill}=\frac{1}{\bsz}\sum_{i=1}^{\bsz}\sum_{j=1}^{\seqlen}(\mathbf{H}_{full}^{(i)}[j] - \mathbf{H}_{\ourname}^{(i)}[j])^2
\end{equation}
where $\bsz$ is the Batch Size. We only use the context information from the real-world dataset to ensure that the token information does not rely on specific prompts. However, the synthetic dataset might contain duplicated contexts, and hence, we only optimize the synthetic dataset with its corresponding labels~\cite{xiao-arxiv24a}. Additionally, we control the sparsity with the regularization value $\SparseRegularizedPar$ instead of the additional loss item defined in DuoAttention, and therefore, only optimize \ourname{} with $\mathcal{L}_{distill}$ as the loss function. We fine-tune two long-context models (Llama-3.1-8B-Instruct~\cite{grattafiori-arxiv24a1} and Mistral-7B-v0.3-Instruct~\cite{jiang-arxiv23a}) on two Nvidia H100 PCIe GPUs for one epoch using AdamW~\cite{kingma-iclr15a, loshchilov-iclr19a} with a learning rate of 2e-3 with a warm-up from 2e-4 in the first 20\% steps and reducing back to 2e-4 in the last 20\% steps. We apply different $\SparseRegularizedPar$ to allow for different sparsity. For a fair comparison, we set the sliding window size \SlidingWindowSize{} as 256, the same as DuoAttention. 

In addition to the baselines in Section~\ref{sec:trainfromscratch}, we also evaluate other state-of-the-art KV eviction strategies, including SnapKV~\cite{li-arxiv24a}, AdaKV~\cite{feng-arxiv24a}, ChunkKV~\cite{liu-arxiv25a}, PyramidKV~\cite{cai-arxiv24a}, CriticalKV~\cite{feng-arxiv25a}. These approaches rely on a sliding window from the query matrix during the pre-filling stage to identify the important tokens. Unlike \ourname{} and the other approaches listed above that update the KV caches on the fly, these approaches only evict the KV cache once after the prefilling stage. Additionally, we evaluate DuoAttention~\cite{xiao-arxiv24a} and MoA~\cite{fu-arxiv24a}. Both approaches extend the existing streaming LLMs approach to provide different budgets for different attention heads. These two methods share the same idea as \ourname{}, where the importance of different KV caches should be learned instead of the pre-defined rules. However, instead of learning the importance of the head level, we aim at learning this information directly on the token level. 

Since we only optimize the \AttScoreLayer{}, the number of learnable parameters is $n_{layers} \cdot d_{model} \cdot n_{heads} \cdot  {n_{options}}$, where $n_{options}$ is the number of options in our search space (in our case, it is $3$). Hence, the size of the parameters that need to be stored is negligible (i.e., it only takes roughly $13$MB on disk for each set of \AttScoreLayer{}) compared to the weights of the LLM. Hence, users could pre-collect all these weights and apply the one that fits their compression requirement.

We evaluate all the baseline results on the RULER~\cite{hsieh-arxiv24a} and LongBench~\cite{bai-acl24a}. RULER is a synthetic benchmark that evaluates a model's long-context capabilities across 4 task categories. LongBench evaluates the model's capacity to understand information in different long context scenarios. Following the setting from KV Press~\cite{kvress-github24a}, we ask all the one-time KV eviction approaches to evict the KV cache once the model receives all the context information before the question-related information arrives. For the other approaches, we evict the KV caches dynamically when new tokens are generated. 

% \begin{wraptable}[17]{r}{11cm}
% \vspace{-3.5mm}
%     \scalebox{0.725}{\input{tables/ruler/llama3-50}}
%     \scalebox{0.725}{\input{tables/ruler/llama3-25}}
%         \caption{Ruler results with 50\% KV budgets (top) and 25\% KV budget (bottom) on LLama 3.1. All Full models use the 100\% KV budgets. We mark the actual KV budgets used by \ourname{} in the bracket. \label{tab:rulerllama}}
% \end{wraptable}

\begin{table}[h]%[17]{r}{11cm}
%\vspace{-3.5mm}
\centering
    \scalebox{0.8}{\begin{tabular}{l|l|llllllllll}
\toprule
 & Full & Duo & SLLM & H2O & Snap & Ada & Chunk & Critical & Pyradmid & MoA & NAtS \\
\midrule
4k & 95.06 & 94.25 & 59.37 & 83.03 & 54.90 & 70.03 & 69.19 & 87.52 & 56.41 & 44.63 & \textbf {95.32} (51\%) \\
8k & 93.22 & 92.42 & 48.58 & 81.51 & 54.10 & 71.43 & 67.20 & 89.01 & 54.55 & 19.80 & \textbf {93.30} (45\%) \\
16k & 90.53 & 88.24 & 50.97 & 65.55 & 54.38 & 76.32 & 67.07 & 88.03 & 58.17 & 15.16 & \textbf {90.01} (42\%) \\
32k & 85.95 & 84.65 & 45.68 & 11.48 & 47.13 & 72.48 & 55.88 & 82.88 & 48.05 & 12.79 & \textbf {86.16} (41\%) \\
64k & 84.01 & 83.54 & 51.40 & 0.07 & 69.61 & 78.10 & 71.05 & 81.93 & 70.49 & 10.86 & \textbf {84.09} (41\%) \\
128k & 73.83 & 73.11 & 46.95 & 0.00 & 39.26 & 64.49 & 41.61 & 70.53 & 40.02 & 11.05 & \textbf {74.77} (44\%) \\
\bottomrule
\end{tabular}
}
    \scalebox{0.8}{\begin{tabular}{l|l|llllllllll}
\toprule
 & Full & Duo & SLLM & H2O & Snap & Ada & Chunk & Critical & Pyradmid & MoA & NAtS \\
\midrule
4k & 95.06 & 73.18 & 37.30 & 71.95 & 35.67 & 44.34 & 51.59 & 57.73 & 35.60 & 35.60 & \textbf {93.93} (24\%) \\
8k & 93.22 & 62.83 & 28.29 & 65.34 & 33.92 & 41.46 & 45.87 & 62.83 & 35.17 & 25.69 & \textbf {90.29} (19\%) \\
16k & 90.53 & 52.01 & 28.19 & 47.82 & 32.74 & 41.61 & 49.40 & 67.08 & 34.35 & 23.94 & \textbf {88.87} (16\%) \\
32k & 85.95 & 53.50 & 30.54 & 9.71 & 32.17 & 40.24 & 39.19 & 64.41 & 32.49 & 20.17 & \textbf {84.40} (15\%) \\
64k & 84.01 & 65.57 & 31.42 & 0.04 & 57.36 & 65.98 & 57.01 & 70.85 & 58.02 & 19.29 & \textbf {79.36} (15\%) \\
128k & 73.83 & 53.05 & 31.41 & 0.00 & 27.36 & 34.33 & 37.35 & 56.51 & 27.34 & 13.40 & \textbf {64.26} (16\%) \\
\bottomrule
\end{tabular}
}
        \caption{Ruler results with 50\% KV budgets (top) and 25\% KV budget (bottom) on LLama 3.1. All Full models use the 100\% KV budgets. We mark the actual KV budgets used by \ourname{} in the bracket. Models with the
        best performance besides the base Full Attention Models are bold~\label{tab:rulerllama}}
\end{table}

The results on the Ruler Benchmark with LLama 3.1 are shown in Table~\ref{tab:rulerllama}. 
\ourname{} uses  $\SparseRegularizedPar=3e-7$  for the 25\% KV size level and $\SparseRegularizedPar=5e-8$ for the 50\% KV size level. \ourname{} achieves nearly lossless scores across different input context lengths. When the KV budget drops to 25\%,  \ourname{} achieves 96\% of the full attention scores with much smaller KV budget sizes, while the other baselines could no longer keep their performance with the decreased KV budgets. Results with Mistral-7B-Instruct-v0.3 can be found under Appendix~\ref{sec:appres}.

%\subsubsection{Needle in a Haystack}

%For the Needle In a Haystack task, we test all models with a context length ranging from $1\,000$ to $32\,000$ with the needle positions ranging from $0$ to $1$. We assign a $25\%$ KV cache budget to each of the baseline approaches. For \ourname{}, we choose the model with the compression rate closest to $25\%$, which corresponds to a model trained with $\SparseRegularizedPar=1e-7$ and an average KV budget size of $15\%$.

%The results on Needle in a Haystack test are shown in Figure~\ref{fig:expniah}. DuoAttention, H2O, and Streaming LLM all maintain a sliding window with a fixed size. Therefore, they could well capture the needles inserted at the text's tail. However, all the baselines struggle with the correct answers as the needle goes into the middle of the context. In contrast, ~\ourname{} efficiently recognizes the necessary tokens across different layers, heads and positions to preserve only tokens that contain the important information. 

%We also plot the KV budget percentage on the bottom right of Figure~\ref{fig:expniah}. Since we set our Sliding window size \SlidingWindowSize{} as 256, the KV budget for the first 1000 tokens is roughly $30\%$. However, as the context length increases, the compression rates gradually decrease to the over compression rate of $15\%$. This shows \ourname{} could save the important information uniformly across the context in different positions. 

%\subsubsection{LongBench}
\begin{table}[]
    \centering
    \scalebox{0.75}{\begin{tabular}{l|r|llllrrllrl}
\toprule
 & Full & Duo & SLLM & H2O & Snap & Ada & Chunk & Critical & Pyradmid & MoA & NAtS \\
\midrule
NarrativeQA & 31.35 & 22.62 & 27.83 & 21.07 & 27.08 & 29.02 & 26.27 & 30.48 & 28.02 & 19.19 & \textbf {30.53}(13\%) \\
Qasper & 24.73 & 18.99 & 14.53 & 15.00 & 14.24 & 14.84 & 12.71 & 15.79 & 13.43 & 18.57 & \textbf {23.76}(27\%) \\
MultiFieldQA-en & 29.46 & 25.36 & 16.47 & 18.24 & 17.72 & 19.31 & 17.44 & 21.74 & 17.72 & 14.75 & \textbf {28.94}(24\%) \\
MultiFieldQA-zh & 60.01 & 54.95 & 33.94 & 29.14 & 33.77 & 35.34 & 33.49 & 36.30 & 32.76 & 21.23 & \textbf {61.18}(22\%) \\
HotpotQA & 17.06 & 16.06 & 13.95 & 16.47 & 15.58 & 16.12 & 16.50 & \textbf {16.73} & 14.23 & 10.18 & 16.06(20\%) \\
2WikiQA & 16.64 & 16.18 & 13.53 & 11.28 & 13.45 & 13.17 & 13.41 & 13.89 & 12.36 & 11.42 & \textbf {17.05}(23\%) \\
Musique & 11.59 & 8.41 & 8.96 & 11.13 & 10.81 & 10.68 & 11.05 & 11.25 & 9.40 & 5.93 & \textbf {11.42}(18\%) \\
DuReader (zh) & 35.56 & 32.18 & 18.64 & 24.77 & 23.57 & 24.57 & 24.24 & 24.68 & 23.56 & 22.33 & \textbf {34.28}(17\%) \\
GovReport & 34.30 & 27.63 & 28.20 & 26.90 & 28.51 & 28.28 & 28.86 & 29.58 & 27.53 & 25.24 & \textbf {33.82}(19\%) \\
QMSum & 23.30 & 22.28 & 20.67 & 20.39 & 21.41 & 22.05 & 21.27 & 22.27 & 21.98 & 19.59 & \textbf {23.11}(17\%) \\
MultiNews & 27.13 & 24.58 & 21.76 & 22.80 & 23.14 & 23.50 & 22.72 & 24.03 & 23.00 & 23.30 & \textbf {26.72}(33\%) \\
VCSUM (zh) & 16.36 & 14.97 & 14.26 & 14.85 & 14.80 & 15.18 & 14.92 & 15.46 & 15.21 & 14.58 & \textbf {15.61}(14\%) \\
TREC & 72.50 & 58.50 & 66.00 & 55.00 & 53.50 & 59.00 & 54.50 & 62.00 & 48.50 & 60.50 & \textbf {72.00}(26\%) \\
TriviaQA & 91.15 & 82.81 & 90.69 & 89.98 & 91.00 & 91.55 & 90.72 & 90.97 & 90.92 & 71.51 & \textbf {91.61}(22\%) \\
SAMSum & 43.72 & 40.14 & 41.97 & 41.98 & 43.59 & 43.11 & 42.32 & 43.79 & 43.15 & 42.30 & \textbf {44.21}(16\%) \\
LSHT & 46.50 & 35.00 & 35.50 & 27.50 & 45.00 & 44.00 & 43.50 & 46.00 & 45.00 & 22.00 & \textbf {47.50}(16\%) \\
Passage Count & 6.63 & 4.50 & \textbf {7.22} & 3.40 & 5.52 & 5.14 & 4.11 & 5.90 & 4.70 & 3.79 & 7.12(20\%) \\
PassageRetrieval-en & 97.98 & 91.65 & 94.81 & 72.25 & 90.36 & 90.70 & 90.38 & 94.70 & 88.32 & 28.21 & \textbf {95.81}(19\%) \\
PassageRetrieval-zh & 77.99 & 55.22 & 28.97 & 52.13 & 73.06 & 78.74 & 55.95 & \textbf {80.15} & 74.66 & 24.78 & 78.5(20\%) \\
LCC & 54.10 & 54.27 & 52.38 & 54.52 & \textbf {55.40} & 54.93 & 54.12 & 54.26 & 54.90 & 54.14 & 53.54(38\%) \\
RepoBench-P & 51.39 & \textbf {57.26} & 49.65 & 55.54 & 51.80 & 52.44 & 52.87 & 52.57 & 52.41 & 53.35 & 53.48(28\%) \\
\bottomrule
\end{tabular}

}
    \caption{LongBench Results with 25\% Budget Allocation. The Full model uses the 100\% KV budgets. The numbers in the brackets for \ourname{} are the used value KV cache sizes. We bold the models with the best performance besides the base Full Attention Models~\label{tab:ResLLBlama25}.}
\end{table}
For the LongBench tasks, we evaluate all the baselines with $50\%$ and $25\%$ of the KV cache sizes. We show the result with KV cache size of $25\%$ for LLama3.1-8B in Table~\ref{tab:ResLLBlama25}. The model that is used in this task is a model with $\SparseRegularizedPar=3e-7$. \ourname{} achieves the best performance on most of the datasets with (in many cases) smaller KV cache values.
We provide further results, including results with Mistral-7B and ~\ourname{} trained with other parameters, in Appendix ~\ref{sec:appres}. 

\subsection{Latency Evaluation}
We evaluate the latency and memory usage with the full attention Llama 3.1 with ~\ourname{} (a $\SparseRegularizedPar=3e-7$) during both pre-filling and decoding phases with different context lengths. The experiments are implemented on a single Nvidia H100 PCIe GPU, and the model is stored as Bfloat16. We adapted RoPE~\citep{su-nc24a} and RMSNorm kernel from FlashInfer~\citep{ye-arxiv25a} to accelerate the forward process. We apply chunked-prefilling ~\cite{agrawal-arxiv23a, Kwon-sosp23a} with a chunk size of $10\,000$ by dividing the input contexts into multiple smaller chunks to reduce the peak memory usage from the intermediate activation values in the FFN layers. The full attention transformer quickly runs out of the GPU memory with a context length of around $200\,000$, while \ourname{} could efficiently reduce the pre-filling and decoding latency and memory usage, allowing \ourname{} to do inference to a context length up to $700\,000$, $3.5$ times larger than the full attention transformer. For the context length of $200\,000$, ~\ourname{} consumes $2.24\times$ less memory with a $3.0\times$ latency speed up during pre-filling and $2.6\times$ less memory with a $1.38~\times$ decoding speed up.

\begin{figure}
    \centering
    \includegraphics[width=0.45\linewidth]{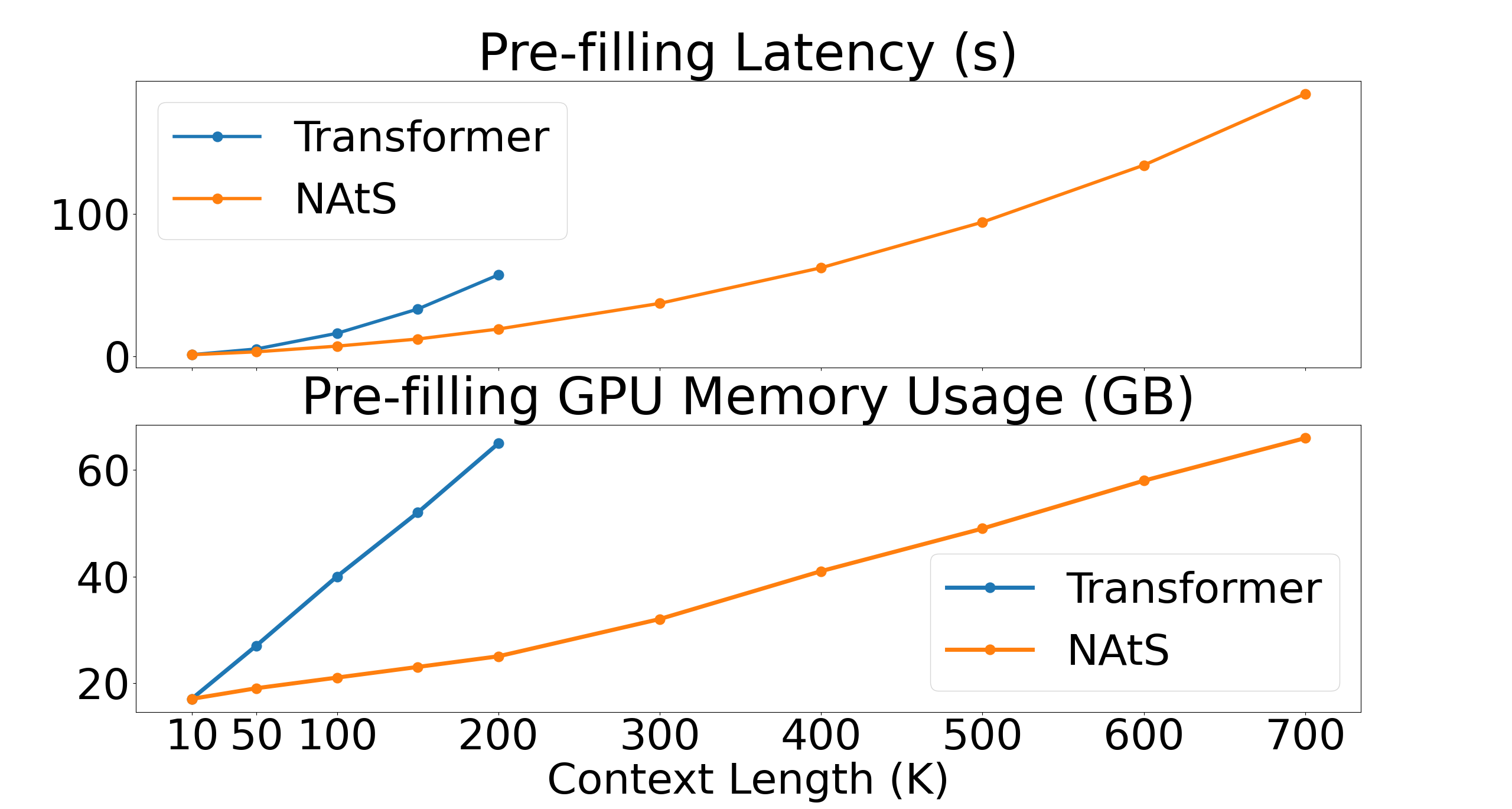}
        \includegraphics[width=0.45\linewidth]{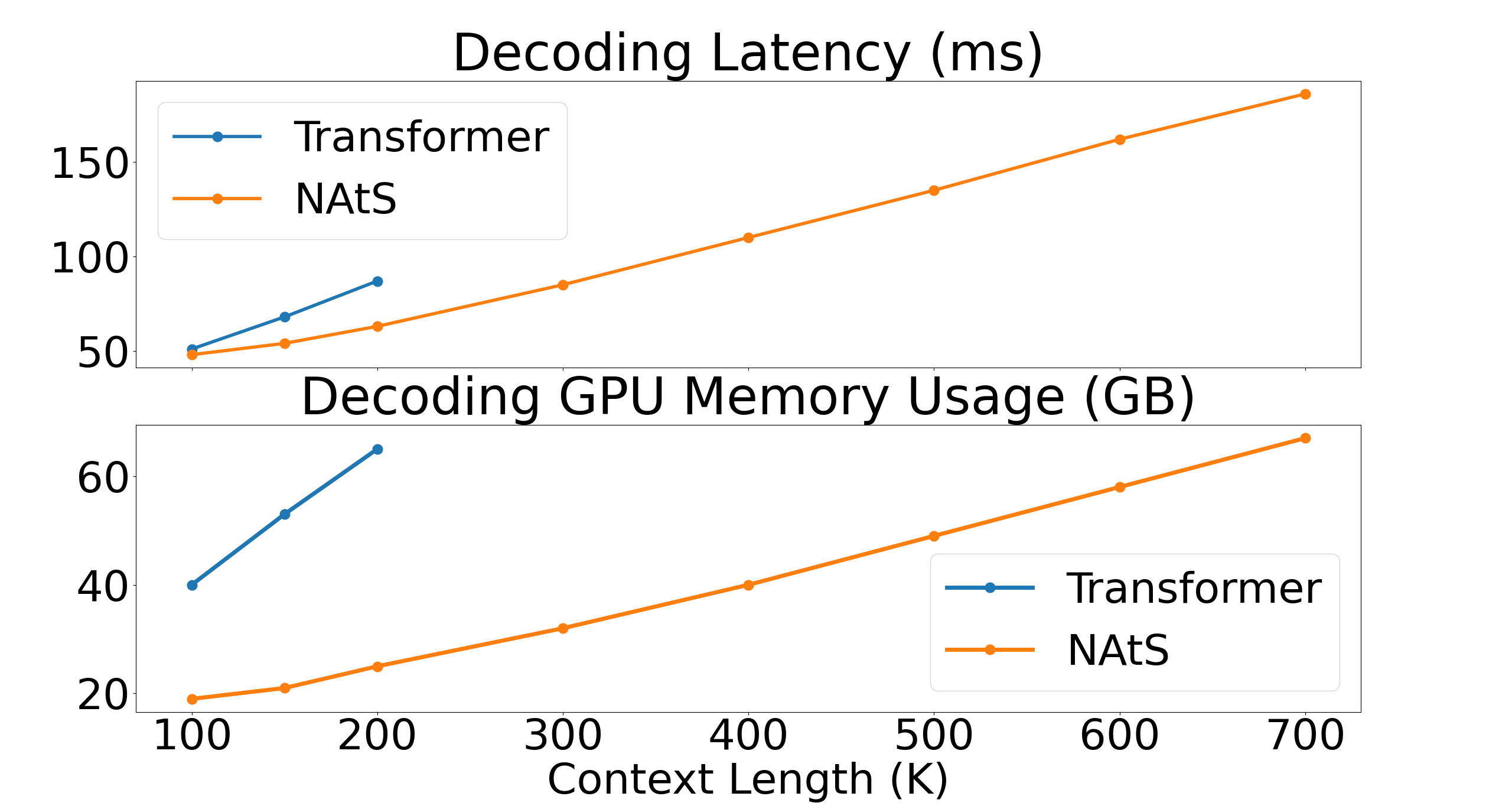}
    \caption{Memory and latency usage during pre-filling (left) and decoding (right)}
    \label{fig:enter-label}
\end{figure}

% \subsubsection{KV Size Distribution~\label{sec:KVSizeDistribution}}
% \begin{figure}
% \centering
%     \subfigure{\includegraphics[width=0.5\textwidth]{figures/Exp/KVSizes/llama/kv_size_Llama_1e7_narrativeqa.png}}
%     \subfigure{\includegraphics[width=0.5\textwidth]{figures/Exp/KVSizes/llama/kv_size_Llama_1e7_repobench-p.png}}
%     \caption{KV cache size compression on different layers~\label{fig:KVSizeDistributions}}
% \end{figure}
% Figure~\ref{fig:KVSizeDistributions} shows the distribution of different KV budgets on the narrativeQA~\cite{kocisky-tacl18a} and RepoBench-p~\cite{liu-iclr24b} dataset. The overall trends are similar: The first few layers and some specific heads require more layers than the other heads. However, as the model goes deeper, the model tends to drop more tokens for the NarrativeQA dataset while still maintaining many tokens for RepoBench-P, especially after layer 16. This shows that a fixed rule might not always fit all the sequences and there is a need to adjust the sampling strategy with different inputs. 

\section{Conclusion and Future Work}
Efficiently managing the KV cache is crucial for deploying large language models in long-context applications.
In this work, we propose \ourname{}, an approach that automatically optimizes the optimal roles of each token to determine how long they should survive. By constructing a learnable mask, \ourname{} learns to sparsify the attention map end-to-end. Our experiments show that \ourname{} uses much less KV caches compared to the State-of-the-art KV caching reduction approach. 
While \ourname{} shows promising results, future work could include exploration of further token roles and structured search spaces with hierarchies. Overall, we believe that \ourname{} paves the way towards more efficient and scalable inference with LLMs.
%However, there are still many hyperparameters that have a strong impact on our approach, such as the sliding window size \SlidingWindowSize{} and the sparse regularization term $\SparseRegularizedPar$. Additionally, the search space proposed in our paper might be limited. One future direction is to propose a larger search space that allows for constructing a more sophisticated sparse attention map. 

\section{Acknowledgement}
The authors gratefully acknowledge the computing time provided to them on the high-performance computers Noctua2 at the NHR Center PC2 under the project hpc-prf-intexml. These are funded by the Federal Ministry of Education and Research and the state governments participating on the basis of the resolutions of the GWK for the national high performance computing at universities (www.nhr-verein.de/unsere-partner).

Difan Deng was supported by the Federal Ministry of Education and Research (BMBF) under the project AI service center KISSKI (grantno.01IS22093C).

\bibliography{bibtex/strings,  bibtex/lib, bibtex/bibtex_local, bibtex/proc, bibtex/proc_local}
\bibliographystyle{plainnat}
%%%%%%%%%%%%%%%%%%%%%%%%%%%%%%%%%%%%%%%%%%%%%%%%%%%%%%%%%%%%

\newpage
\section*{NeurIPS Paper Checklist}

%%% BEGIN INSTRUCTIONS %%%
The checklist is designed to encourage best practices for responsible machine learning research, addressing issues of reproducibility, transparency, research ethics, and societal impact. Do not remove the checklist: {\bf The papers not including the checklist will be desk rejected.} The checklist should follow the references and follow the (optional) supplemental material.  The checklist does NOT count towards the page
limit. 

Please read the checklist guidelines carefully for information on how to answer these questions. For each question in the checklist:
\begin{itemize}
    \item You should answer \answerYes{}, \answerNo{}, or \answerNA{}.
    \item \answerNA{} means either that the question is Not Applicable for that particular paper or the relevant information is Not Available.
    \item Please provide a short (1–2 sentence) justification right after your answer (even for NA). 
   % \item {\bf The papers not including the checklist will be desk rejected.}
\end{itemize}

{\bf The checklist answers are an integral part of your paper submission.} They are visible to the reviewers, area chairs, senior area chairs, and ethics reviewers. You will be asked to also include it (after eventual revisions) with the final version of your paper, and its final version will be published with the paper.

The reviewers of your paper will be asked to use the checklist as one of the factors in their evaluation. While "\answerYes{}" is generally preferable to "\answerNo{}", it is perfectly acceptable to answer "\answerNo{}" provided a proper justification is given (e.g., "error bars are not reported because it would be too computationally expensive" or "we were unable to find the license for the dataset we used"). In general, answering "\answerNo{}" or "\answerNA{}" is not grounds for rejection. While the questions are phrased in a binary way, we acknowledge that the true answer is often more nuanced, so please just use your best judgment and write a justification to elaborate. All supporting evidence can appear either in the main paper or the supplemental material, provided in the appendix. If you answer \answerYes{} to a question, in the justification please point to the section(s) where related material for the question can be found.

%%% END INSTRUCTIONS %%%

\begin{enumerate}
\item {\bf Claims}
    \item[] Question: Do the main claims made in the abstract and introduction accurately reflect the paper's contributions and scope?
    \item[] Answer: \answerYes{} % Replace by \answerYes{}, \answerNo{}, or \answerNA{}.
    \item[] Justification: In this paper, we present \ourname{} in Section~\ref{sec:approach}. The experimental results in Section~\ref{sec:exp} show that its strong performance and efficiency in different benchmarks.

\item {\bf Limitations}
    \item[] Question: Does the paper discuss the limitations of the work performed by the authors?
    \item[] Answer: \answerYes{} % Replace by \answerYes{}, \answerNo{}, or \answerNA{}.
    \item[] Justification: Yes, we discuss the limitations in Section ~\ref{sec:limitation}. 

\item {\bf Theory assumptions and proofs}
    \item[] Question: For each theoretical result, does the paper provide the full set of assumptions and a complete (and correct) proof?
    \item[] Answer: \answerNA{} % Replace by \answerYes{}, \answerNo{}, or \answerNA{}.
    \item[] Justification: This paper does not include theoretical results

    \item {\bf Experimental result reproducibility}
    \item[] Question: Does the paper fully disclose all the information needed to reproduce the main experimental results of the paper to the extent that it affects the main claims and/or conclusions of the paper (regardless of whether the code and data are provided or not)?
    \item[] Answer: \answerYes{} % Replace by \answerYes{}, \answerNo{}, or \answerNA{}.
    \item[] Justification: The detailed experiments set up and hyperparameter configurations are described in Section ~\ref{sec:exp} and ~\ref{sec:expdetails}. The search space is described in Section~\ref{sec:approach} and the algorithm is described in Section~\ref{sec:nats-fa2}. We also provide the codes under \url{https://github.com/automl/NeuralAttentionSearch}

\item {\bf Open access to data and code}
    \item[] Question: Does the paper provide open access to the data and code, with sufficient instructions to faithfully reproduce the main experimental results, as described in supplemental material?
    \item[] Answer: \answerYes{} % Replace by \answerYes{}, \answerNo{}, or \answerNA{}.
    \item[] Justification: Yes, all the datasets used in this paper are open-sourced. We further provide a script on how to generate the training set for fine-tuning an LLM with \ourname{} under: \url{https://github.com/automl/NeuralAttentionSearch}

\item {\bf Experimental setting/details}
    \item[] Question: Does the paper specify all the training and test details (e.g., data splits, hyperparameters, how they were chosen, type of optimizer, etc.) necessary to understand the results?
    \item[] Answer: \answerYes{} % Replace by \answerYes{}, \answerNo{}, or \answerNA{}.
    \item[] Justification: Yes, we provide the detailed experiments setup and hyperparemter configurations in Section~\ref{sec:exp} and ~\ref{sec:expdetails}

\item {\bf Experiment statistical significance}
    \item[] Question: Does the paper report error bars suitably and correctly defined or other appropriate information about the statistical significance of the experiments?
    \item[] Answer: \answerNo{} % Replace by \answerYes{}, \answerNo{}, or \answerNA{}.
    \item[] Justification:  All the experiments in this paper run only once with a fixed random seed.

\item {\bf Experiments compute resources}
    \item[] Question: For each experiment, does the paper provide sufficient information on the computer resources (type of compute workers, memory, time of execution) needed to reproduce the experiments?
    \item[] Answer: \answerYes{} % Replace by \answerYes{}, \answerNo{}, or \answerNA{}.
    \item[] Justification: Yes, we report the hardware and resources we used to train (4 Nvidia H100 PCIe GPUs for 18 hours) /fine-tune the model (2 Nvidia H100 PCIe GPUs in 8 hours) in Section~\ref{sec:exp} and ~\ref{sec:expdetails}. 
    
\item {\bf Code of ethics}
    \item[] Question: Does the research conducted in the paper conform, in every respect, with the NeurIPS Code of Ethics \url{https://neurips.cc/public/EthicsGuidelines}?
    \item[] Answer: \answerYes{} % Replace by \answerYes{}, \answerNo{}, or \answerNA{}.
    \item[] Justification: Yes, we have reviewed the NeurIPS Code of Ethics. ~\ourname{} is a methodology paper and does not have ethical concerns. 

\item {\bf Broader impacts}
    \item[] Question: Does the paper discuss both potential positive societal impacts and negative societal impacts of the work performed?
    \item[] Answer: \answerYes{}{} % Replace by \answerYes{}, \answerNo{}, or \answerNA{}.
    \item[] Justification: Yes, we have discussed the broader impact in Section~\ref{sec:impacts}
    
\item {\bf Safeguards}
    \item[] Question: Does the paper describe safeguards that have been put in place for responsible release of data or models that have a high risk for misuse (e.g., pretrained language models, image generators, or scraped datasets)?
    \item[] Answer: \answerNA{} % Replace by \answerYes{}, \answerNo{}, or \answerNA{}.
    \item[] Justification: This paper poses no such risks. 

\item {\bf Licenses for existing assets}
    \item[] Question: Are the creators or original owners of assets (e.g., code, data, models), used in the paper, properly credited and are the license and terms of use explicitly mentioned and properly respected?
    \item[] Answer: \answerYes{} % Replace by \answerYes{}, \answerNo{}, or \answerNA{}.
    \item[] Justification: Yes, the data set and models used in our paper are open sourced and we cite the original paper/github that provides the data and model in this paper. 

\item {\bf New assets}
    \item[] Question: Are new assets introduced in the paper well documented and is the documentation provided alongside the assets?
    \item[] Answer: \answerNA{} % Replace by \answerYes{}, \answerNo{}, or \answerNA{}.
    \item[] Justification: This paper does not release new assets

\item {\bf Crowdsourcing and research with human subjects}
    \item[] Question: For crowdsourcing experiments and research with human subjects, does the paper include the full text of instructions given to participants and screenshots, if applicable, as well as details about compensation (if any)? 
    \item[] Answer: \answerNA{} % Replace by \answerYes{}, \answerNo{}, or \answerNA{}.
    \item[] Justification: This paper does not involve crowdsourcing nor research with human subjects. 

\item {\bf Institutional review board (IRB) approvals or equivalent for research with human subjects}
    \item[] Question: Does the paper describe potential risks incurred by study participants, whether such risks were disclosed to the subjects, and whether Institutional Review Board (IRB) approvals (or an equivalent approval/review based on the requirements of your country or institution) were obtained?
    \item[] Answer: \answerNA{} % Replace by \answerYes{}, \answerNo{}, or \answerNA{}.
    \item[] Justification: This paper does not involve crowdsourcing nor research with human subjects

\item {\bf Declaration of LLM usage}
    \item[] Question: Does the paper describe the usage of LLMs if it is an important, original, or non-standard component of the core methods in this research? Note that if the LLM is used only for writing, editing, or formatting purposes and does not impact the core methodology, scientific rigorousness, or originality of the research, declaration is not required.
    %this research? 
    \item[] Answer: \answerYes{} % Replace by \answerYes{}, \answerNo{}, or \answerNA{}.
    \item[] Justification: The experiments parts involve fine tuning an existing LLM model. We used ~\ourname{} to fine tune a Meta-Llama-3.1-8b-Instruct and Mistral-7B-Instruct-v0.3 model. 

\end{enumerate}

\newpage
\appendix

\section{Limitations~\label{sec:limitation}}
While \ourname{} shows promising results that efficiently reduce the KV cache sizes and our search space already contains most of the KV cache eviction strategies. However, the construction of the attention mask is still column-wise oriented and mostly focuses on the vertical style sparse attention~\cite{jiang-neurips24a}. Hence, this work does not involve the block-sparse attention and slash attention that are widely observed 
in other frameworks~\cite{jiang-neurips24a, lai-iclr25a}. Which in turn might result in a more sophisticated searching and inference process, as we can no longer easily drop the desired tokens since they can also be related to the value of the Q matrix. One potential future direction would be to introduce a more structured search space that allows for the construction of more sparse attention types. 

%However, there are still many hyperparameters that might impact ~\ourname{}, such as the sliding window size \SlidingWindowSize{} and the sparse regularization term $\SparseRegularizedPar$. Additionally, the search space proposed in our paper might be limited. One future direction is to propose a larger search space that allows for constructing a more sophisticated sparse attention map.  

\section{Broader Impact~\label{sec:impacts}}
LLMs are widely applied to different fields nowadays. However, the cost for LLM to store the KV cache and predict the next token is still huge, given the $\mathcal{O}(\seqlen)$ computation and memory costs of full Attention Models. This prevents further adaptation of LLMs (and other transformer-based foundation models with properties similar to those described before) because of high energy consumption and limited context windows. \ourname{} achieves a substantial reduction in KV cache size with minimal impact on model performance, outperforming existing state-of-the-art approaches. This increased efficiency can enable the deployment of larger, more powerful language models on resource-constrained devices and facilitate the development of new applications that rely on long-context understanding, such as advanced conversational AI, comprehensive document summarization, and complex code generation. By making long-context processing more accessible, \ourname{} has the potential to accelerate progress in natural language processing and related fields. Nevertheless, it does not solve other inherent problems of LLMs such as hallucinations.

\section{Details on Backward Propagation~\label{apd:backward}}
\subsection{Gradients for Attention Masks}

To compute the gradients for \attMSK, we set $g(\attMap, \attMSK) = e^{\attMap} \odot \attMSK$; then the gradient for \attMSK{} is: 
\begin{align}
   \frac{\partial \attO}{\partial \attMSK} &= 
   \frac{\partial \attO}{\partial g}\frac{\partial g}{\partial \attMSK} \\
   \frac{\partial g}{\partial \attMSK} &= e^{\attMap}\label{eq:gradientMSK} \\
   \frac{\partial g}{\partial \attMap} &= e^{\attMap} \odot \attMSK\label{eq:gradientA}
\end{align}

In Eq.~\ref{eq:gradientMSK} and ~\ref{eq:gradientA}, we show the gradient for $\attMSK$ is the same as the value that $\partial \attO / \partial \attMap$ is supposed to be if no mask is applied. 
Since we have $g_{i,j}=e^{\attMap_{i,j}} \odot \attMSK_{i, j}$. Let's set $S_{i}:=\sum_j g_{i,j}$ and $P_{i,j}:=\frac{g_{i,j}}{S_{i}}$. Then we have $\mathbf{d}\mathbf{P} = \mathbf{d}\mathbf{\attO}\mathbf{V}^T$. Therefore, 

\begin{equation}
   dg_{i:} = (diag(\frac{1}{S_{i:}}) - \frac{1}{S_{i:}}P_{i:}^T)dP_{i:}\label{eq:gradGi}
\end{equation}
Combining Eq.~\ref{eq:gradGi} with Eq.~\ref{eq:gradientA} and Eq.~\ref{eq:gradientMSK} provides the same gradients as the vanilla softmax function with additive masks:
\begin{align}
    d\attMSK_{i:} &= (diag(\frac{e^{\attMap_{i:}}}{S_{i:}}) - \frac{e^{\attMap_{i:}}}{S_{i:}}P_{i:}^T)dP_{i:}\\
    d\attMap &= d\attMSK \odot  \attMSK  \nonumber \\   
             &= (diag(P_{i:}) - P_{i:}P^T_{i:})dP_{i:}
    \label{eq:gradientA1}
\end{align}

Since $dg_{i:}$ is required to compute the gradient for $\mathbf{A}_{i:}$ and always needs to be computed. We can directly use this information to compute the gradients for the attention Mask $\attMSK$.

Following FlashAttention~\cite{dao-neurips22a}, we define $D_{i} = do_i^To_i$, then 
\begin{align}
    d\attMSK_{i,j} = \frac{e^{A_{i,j}}}{S_{i:}}(dP_{ij} - D_{i})~\label{eq:dmsk}\\
    d\attMap_{i,j} = P_{i,j}(dP_{ij} - D_{i})~\label{eq:dattnvalues}
\end{align}

and $d\attMap$ is computed by Eq.~\ref{eq:gradientA1}. After that, we can backpropagate $d\attMap$ to $dq$ and $dv$. Since $\attMSK$ needs to be recomputed anyway in the flash attention's backward process, this only results in little computational overhead. 

However, in practice, we cannot guarantee an upper bound for $\frac{e^{A_{i,j}}}{S_i}$ with $\attMSK_{i,j}=0$ since $S_{i}:=\sum_j e^{\attMap_{i,j}} \odot \attMSK_{i, j}$. As a result, $\lim_{e^{(A_{i,j})}\to \infty}(\frac{e^{(A_{i,j})}}{S_i})= \infty$ when $\attMSK_{i,j}=0$. Hence, we first clip $\frac{e^{A_{i,j}}}{S_i}$ within $(0, 1)$ and then compute $d\attMSK$ with the clipped value:
\begin{align}
    P'_{i,j} &= \min(\frac{e^{A_{i,j}}}{S_i}, 1) \\
    d\attMSK_{i,j} &= P'_{i,j}(dP_{ij} - D_{i})~\label{eq:dmskclip}
\end{align}

\subsection{Details on computing the gradients for computing the token gradients info $d\tokenstates$}
The gradients towards each token are collected through the column sum of each value weighted by the corresponding attention masks:
\begin{equation}
    d \tokenstates_{i} = \sum_{j} d \attMSK_{i,j} \times  \attMSK_{i,j}^{\tokenstates} ~\label{eq:gradmsk}
\end{equation}
Where $\tokenstates \in \{G, L, SW\}$ is the discretized token type. Intuitively, this shows the model's preference over short-range or long-range correlations: If $i$ is quite close to $j$, then all the $\tokenstates$ will receive the same gradient information. However, if the model wants to create a long-range correlation with $i \gg j$, only \GlobalTokens{} will receive the gradient information. The network will therefore prefer to classify the corresponding tokens as global tokens.

Eq.~\ref{eq:msklocal} shows that the \GlobalToken{} controls the local mask size. Therefore, the gradients for \GlobalToken{} $i$ should also be regularized by the gradient information from the previous tokens:
\begin{equation}
    \frac{\partial \attMSK_{i,j}^{L}}{\partial \EndSeqHard_k} = - \prod_{\substack{n=j+1 \\ n \neq k}}^{i-1}(1-\EndSeqHard_{n}) ~\label{eq:gradlocalmsk}
\end{equation}
In cases where $\EndSeqHard_k$ is 0, Equation~\ref{eq:gradlocalmsk} is the negative value of Equation~\ref{eq:msklocal}. However, for the case where $\EndSeqHard_k = 1$, this is equivalent to the local mask values where $k$ is no longer set as a \GlobalToken. This requires us to find the index of the next global token $\EndSeqHardTrue_{k+1}$ and the last global token $\EndSeqHardTrue_{k-l}$ where $l$ is the length of the local sequence that ends at $k$.

This gradient information will then be collected and subtracted from the computed \GlobalTokens{} gradient values:
\begin{align}
        d \tokenstates^{G-}_{i} &=
    \begin{cases}
          \sum\limits_{\substack{m \geq i\\i > n \geq \EndSeqHardTrue_{k-l}}} \attMSK^L_{m,n} \times d\attMSK_{m,n} \ \  &\text{if}\  \EndSeqHard_i = 0, \\
         \sum\limits_{\substack{\EndSeqHardTrue_{i+1} \geq m >=i\\  i > n \geq \EndSeqHardTrue_{k-l} }} d\attMSK_{m,n} \ \  &\text{if}\  \EndSeqHard_i = 1, 
    \end{cases}\label{eq:gradcollected1} \\
    d \tokenstates^{G}_{i} &= d \tokenstates^{G}_{i} - d \tokenstates^{G-}_{i} \label{eq:gradcollected2}
\end{align}

Intuitively, this gradient term $d \tokenstates^{G-}_{i}$ checks if the new \GlobalToken{} needs to be inserted into the current sub-sequence (when $\EndSeqHard$ is 0) or we should remove the current \GlobalToken{} to enlarge the current sub-sequence (when $\EndSeqHard$ is 1). We further illustrate this process in the appendix.

 Figure ~\ref{fig:GradTokenGrad} illustrates an example of this. Token 4 is a global token, and we search for its next \GlobalToken{} (which is Token 10 in this example). Assuming that we want to change Token 4 to another role,  the regions within the boundary (orange ones) are those tokens that are influenced by this swtching: given that Token 4 no longer becomes \GlobalToken{}, Token 2 and 3 will be part of a larger subsequence that ranges from 1 to 10, and the attention maps within the orange region should not be masked out. Intuitively, this gradient measures the regret that we made in order to switch one \GlobalToken{} into a \LocalToken{}. A similar idea can be found in the red region. Assuming that we want to switch Token 7 to a \GlobalToken{}, then Tokens 5,6 will be split into a new subsequence and we will no longer connect them with Tokens 8, 9, 10 since they belong to two different sub-sequences after the switch. Hence, this value in the red regions measures the regret if we mistakenly classify a \GlobalToken{} as non-\GlobalTokens.

\begin{wrapfigure}[11]{r}{0.45\textwidth}
 \vspace{-3mm}
    \centering
    \includegraphics[width=1.0\linewidth]{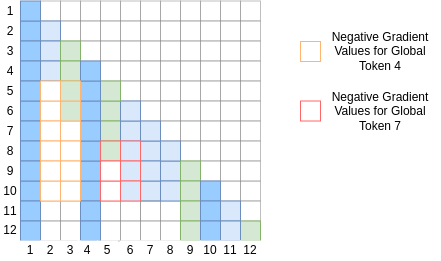}
    \caption{The gradient term $d \tokenstates^{G-}_{i}$ for Token 4 and 7}
    \label{fig:GradTokenGrad}
\end{wrapfigure}

In practice, the $d \tokenstates^{G-}_{i}$ values for \GlobalToken{} can be easily collected by looking at the last and next $\EndSeqHard$ index with a scan function to collect the gradients to the corresponding positions. However, computing $d \tokenstates^{G-}_{i}$ with $G_i =0$ requires the gradient information from all the past token mask information. This might be computationally prohibitive in practice, since the valid values contained in Equation~\ref{eq:gradcollected1} are sparse, we ignore the cases for $\EndSeqHard_i = 0$ and only compute the dense gradients for the case where $\EndSeqHard_i = 1$. 

%However, for those values on non \GlobalTokens, since the entire sequence can be a sub-sequence if no \GlobalToken{} exists in the current sequence, collecting all the values that are below and on the left side of the \LocalTokens{} might be too expensive. Hence, we ask each \LocalToken{} can only visit a fixed number of tokens on both sides. In this case, this operation is equivalent to a convolutional operation with weights of either 1 or 0.  We set this value as 16 in our experiments to fit the grid sizes in triton. 

\subsection{Optimizing for sparser Attention Maps}
The sparse regularization term $\SparseRegularizedPar$ is directly applied to the corresponding gradients for the \GlobalTokens{} and \LocalTokens{}.  Intuitively, if the column-wise sum of the gradients for each iteration is smaller than $\SparseRegularizedPar$, the attention maps of that column would require no further update or gradients updates towards a sparser transformer, they will be push the projection layer to classify those tokens as sliding window tokens. This ensures that the global tokens and local tokens actually require a certain amount of column-wise attention map values to keep them activated. Hence, the sparse regularization term $\SparseRegularizedPar$ could also be considered as a soft threshold for the attention map values, where attention map values smaller than that will be encouraged to be filtered out. This value should therefore penalize the number of unmasked tokens for each row. While the number of unmasked tokens for \GlobalToken{} and \LocalTokens{} in column $i$ are $\seqlen - i$ and $\EndSeqHardTrue_i - i$ with $\EndSeqHardTrue_i$ defined in Section~\ref{sec:natstrain}. Hence, we have:
\begin{align}
    d\EndSeqHard^{G_{sparse}}_i &= \SparseRegularizedPar \times \frac{\seqlen - i}{\seqlen}\label{eq:gradsparseG}\\
    d\EndSeqHard^{L_{sparse}}_i &= \SparseRegularizedPar \times \frac{\EndSeqHardTrue_i - i}{\seqlen}\label{eq:gradsparseI}
\end{align}

Combining Eq. ~\ref{eq:gradmsk}, ~\ref{eq:gradcollected1}, ~\ref{eq:gradsparseG} and Eq. ~\ref{eq:gradsparseI}, we have:
\begin{align}
    d \tokenstates^G_{i} &= \sum_{j} d \attMSK_{i,j} \times  \attMSK_{i,j}^{G}  \times \attMSK_{i,j}^{casual} + d\EndSeqHard^{G_{sparse}}_i - d \tokenstates^{G-}_{i} \label{eq:dmskglobal} \\
    d \tokenstates^L_{i} &= \sum_{j} d \attMSK_{i,j} \times  \attMSK_{i,j}^{L} \times \attMSK_{i,j}^{casual} + d\EndSeqHard^{L_{sparse}}_i \label{eq:dmsklocal} \\
    d \tokenstates^{SW}_{i} &= \sum_{j} d \attMSK_{i,j} \times  \attMSK_{i,j}^{SW} \times \attMSK_{i,j}^{casual} \label{eq:dmsksw}
\end{align}
with $\attMSK_{i,j}^{G}, \attMSK_{i,j}^{SW}, \attMSK_{i,j}^{L}$ defined in Eq. ~\ref{eq:mskglobal}, ~\ref{eq:msklocalfwd} and ~\ref{eq:msksw} and $\attMSK_{i,j}^{casual}$ is a casual attention mask.

\section{Integrating \ourname{} into FlashAttention~\label{sec:nats-fa2}}

\begin{algorithm}[h]
  % \algsetup{linenosize=\tiny}
  \caption{\small\label{alg:nats_fwd_fa2}\ourname{} forward pass on FlashAttention2, \textcolor{blue}{we mark the ~\ourname{} related operations with blue}}
  \begin{algorithmic}[1]
    \REQUIRE Matrices $\mathbf{Q}, \mathbf{K}, \mathbf{V} \in \mathbb{R}^{\seqlen \times d}$, \textcolor{blue}{token states $\mathbf{\tokenstates} \in \mathbb{R}^{\seqlen \times \nopts}$ , indices of the next token states $\mathbf{\EndSeqHardTrue} \in \mathbb{R}^{N}$} in HBM, block sizes $B_c$, $B_r$, \textcolor{blue}{sliding window size $\SlidingWindowSize$}
    \STATE Divide $\mathbf{Q}$ into $T_r = \left\lceil\frac{N}{B_r} \right\rceil$ blocks $\mathbf{Q}_1, \dots, \mathbf{Q}_{T_r}$ of size $B_r \times d$ each,
    and divide $\mathbf{K}, \mathbf{V}$, \textcolor{blue}{$\mathbf{\tokenstates}, \mathbf\EndSeqHardTrue$} into $T_c = \left\lceil \frac{N}{B_c} \right\rceil$ blocks $\mathbf{K}_1, \dots, \mathbf{K}_{T_c}$,
    $\mathbf{V}_1, \dots, \mathbf{V}_{T_c}$, of size $B_c \times d$, \textcolor{blue}{ $\mathbf{\tokenstates}_1, \dots, \mathbf{\tokenstates}_{T_c}$ of size $B_r \times \nopts$ , and
    $\mathbf{\EndSeqHardTrue}_1, \dots, \mathbf{\EndSeqHardTrue_{T_c}}$ of size $B_r$} each.
    
    \STATE Divide the output $\vO \in \mathbb{R}^{N \times d}$ into $T_r$ blocks $\vO_i, \dots, \vO_{T_r}$ of size
    $B_r \times d$ each, and divide the logsumexp $L$ into $T_r$ blocks $L_i, \dots, L_{T_r}$ of size
    $B_r$ each.
    \FOR{$1 \le i \le T_r$} \label{alg:stream_attn_outer_loop}
      \STATE \label{alg:stream_attn_load_q} Load $\vQ_i$ from HBM to on-chip SRAM.
      \STATE \label{alg:stream_attn_init} On chip, initialize $\vO_{i}^{(0)} = (0)_{B_r \times d} \in \mathbb{R}^{B_r \times d}, \ell_{i}^{(0)} = (0)_{B_r} \in \mathbb{R}^{B_r}, m_{i}^{(0)} = (-\infty)_{B_r} \in \mathbb{R}^{B_r}$.
      \FOR{$1 \le j \le T_c$}
        \STATE \textcolor{blue}{Load $\mathbf{\tokenstates}_j$ and $\mathbf{\EndSeqHardTrue}_j$ to SRAM and check if the current block contain any valid values according to Equations~\ref{eq:mskglobal},\ref{eq:msksw},\ref{eq:msklocalfwd},}
        \IF {\textcolor{blue}{Do Compute}}
        \STATE \textcolor{blue}{Construct attention mask $\mathbf{\attMSK}_{i,j}$ with  $\mathbf{\tokenstates}_j$, $\mathbf{\EndSeqHardTrue}_j$, and $\SlidingWindowSize$ with Equations~\ref{eq:mskglobal},\ref{eq:msksw},\ref{eq:msklocalfwd}}
            \STATE \label{alg:stream_attn_load_kv} Load $\vK_j, \vV_j$ from HBM to on-chip SRAM.
            \STATE \label{alg:stream_attn_qk} On chip, compute $\vS_{i}^{(j)} = \vQ_i \vK_j^T \in \mathbb{R}^{B_r \times B_c}$.
            \STATE \label{alg:stream_attn_statistics} On chip, compute
            $m_{i}^{(j)} = \mathrm{max}(m_{i}^{(j-1)}, \mathrm{rowmax}(\vS_{i}^{(j)})) \in \mathbb{R}^{B_r}$, $\tilde{\vP}_{i}^{(j)} = \exp(\vS_{i}^{(j)} - m_{i}^{(j)}) \textcolor{blue}{ \odot \mathbf{\attMSK}_{i,j}} \in \mathbb{R}^{B_r \times B_c}$ (pointwise),
            $\ell_{i}^{(j)} = e^{m_{i}^{j-1} - m_{i}^{(j)}} \ell_{i}^{(j-1)} + \mathrm{row sum}(\tilde{\vP}_{i}^{(j)}) \in \mathbb{R}^{B_r}$.
            \STATE \label{alg:stream_attn_update} On chip, compute
            $\vO_{i}^{(j)} = \diag(e^{m_{i}^{(j-1)} - m_{i}^{(j)}})^{-1} \vO_{i}^{(j-1)} + \tilde{\vP}_{i}^{(j)} \vV_j$.
        \ENDIF
      \ENDFOR
      \STATE On chip, compute $\vO_{i} = \diag(\ell_{i}^{(T_c)})^{-1} \vO_{i}^{(T_c)}$.
      \STATE On chip, compute $L_{i} = m_{i}^{(T_c)} + \log(\ell_i^{(T_c)})$.
      \STATE Write $\vO_{i}$ to HBM as the $i$-th block of $\vO$.
      \STATE Write $L_{i}$ to HBM as the $i$-th block of $L$.
    \ENDFOR
    \STATE Return the output $\vO$ and the logsumexp $L$.
  \end{algorithmic}
\end{algorithm}

Flashattention~\cite{dao-iclr24a} continuously loads Q, K, V values to the SRAM and only computes the attention maps within the loaded blocks. 
~\ourname{} loads the corresponding token states information $\tokenstates$ and construct the masks $\attMSK$ dynamically with Equation~\ref{eq:mskglobal}, ~\ref{eq:msklocal}, ~\ref{eq:msksw}. During the backward process, Flashattention collects the row-wise gradients for computing $d\mathbf{Q}$ and column-wise gradients to compute $d\mathbf{K}$ and $d\mathbf{V}$. Hence, the gradients $d\mathbf{\tokenstates}$ can be computed together with $d\mathbf{K}$ and $d\mathbf{V}$. We show how ~\ourname{} can be integrated into Flashattention2~~\cite{dao-iclr24a} in Algorithms~\ref{alg:nats_fwd_fa2} and \ref{alg:nats_bwd_fa2}. We highlight the difference between the vanilla Flashattention2 and Flashattention2 with ~\ourname{} in blue. Compared with vanilla Flashattention models, ~\ourname{}  adaptively skips the blocks that do not have valid mask values to avoid unnecessary computation. These invalid blocks will only be applied to update the gradients for $~d\mathbf{\tokenstates}$.

\begin{algorithm}[h]
  \caption{\small\label{alg:nats_bwd_fa2}\ourname{} backward pass on FlashAttention2, \textcolor{blue}{we mark the ~\ourname{} related operations with blue}}
  \begin{algorithmic}[1]
    \REQUIRE Matrices $\vQ, \vK, \vV, \vO, \vdO \in \mathbb{R}^{\seqlen \times d}$, \textcolor{blue}{token states $\mathbf{\tokenstates} \in \mathbb{R}^{\seqlen \times \nopts}$ , indices of the next token states $\mathbf{\EndSeqHardTrue} \in \mathbb{R}^{\seqlen}$} in HBM,
    vector $L \in \mathbb{R}^N$ in HBM, block sizes $B_c$, $B_r$, \textcolor{blue}{sliding window size $\SlidingWindowSize$}.
    \STATE Divide $\vQ$ into $T_r = \left\lceil\frac{N}{B_r} \right\rceil$ blocks $\vQ_1, \dots, \vQ_{T_r}$ of size $B_r \times d$ each,
    and divide $\vK, \vV$, \textcolor{blue}{$\mathbf{\tokenstates}, \mathbf\EndSeqHardTrue$} in to $T_c = \left\lceil \frac{N}{B_c} \right\rceil$ blocks $\vK_1, \dots, \vK_{T_c}$ and
    $\vV_1, \dots, \vV_{T_c}$, of size $B_c \times d$, \textcolor{blue}{ $\mathbf{\tokenstates}_1, \dots, \mathbf{\tokenstates}_{T_c}$ of size $B_r \times \nopts$ , and
    $\mathbf{\EndSeqHardTrue}_1, \dots, \mathbf{\EndSeqHardTrue_{T_c}}$ of size $B_r$}  each.
    \STATE Divide $\vO$ into $T_r$ blocks $\vO_i, \dots, \vO_{T_r}$ of size
    $B_r \times d$ each, divide $\vdO$ into $T_r$ blocks $\vdO_i, \dots, \vdO_{T_r}$
    of size $B_r \times d$ each, and divide $L$ into $T_r$ blocks $L_i, \dots, L_{T_r}$ of size
    $B_r$ each.
    \STATE Initialize $\vdQ = (0)_{N \times d}$ in HBM and divide it into $T_r$ blocks $\vdQ_1, \dots, \vdQ_{T_r}$ of size $B_r \times d$ each.
    Divide $\vdK, \vdV \in \mathbb{R}^{N \times d}$ in to $T_c$ blocks $\vdK_1, \dots, \vdK_{T_c}$ and
    $\vdV_1, \dots, \vdV_{T_c}$, of size $B_c \times d$ each. \textcolor{blue}{Divide $d\mathbf{\tokenstates} \in \mathbb{R}^{\nopts}$ in to $d\mathbf{\tokenstates}_1, \dots, d\mathbf{\tokenstates}_{T_c}$ of size $B_r \times \nopts$ each.}
    \STATE Compute $D = \mathrm{rowsum}(\vdO \odot \vO) \in \mathbb{R}^d$ (pointwise multiply), write
    $D$ to HBM and divide it into $T_r$ blocks $D_1, \dots, D_{T_r}$ of size
    $B_r$ each.
    \FOR{$1 \le j \le T_c$}
      \STATE Load $\vK_j, \vV_j$, \textcolor{blue}{$\mathbf{\tokenstates}_j$ and $\mathbf{\EndSeqHardTrue}_j$} from HBM to on-chip SRAM.
      \STATE Initialize $\vdK_j = (0)_{B_c \times d}, \vdV_j = (0)_{B_c \times d}$, \textcolor{blue}{$d\mathbf{\tokenstates}_j = (0)_{B_c \times \nopts}$} on SRAM.
      \FOR{$1 \le i \le T_r$}
        \STATE Load $\vQ_i, \vO_i, \vdO_i, \vdQ_i, L_i, D_i$ from HBM to on-chip SRAM.
        \STATE On chip, compute $\vS_{i}^{(j)} = \vQ_i \vK_j^T \in \mathbb{R}^{B_r \times B_c}$.
        \STATE \textcolor{blue}{On chip, Construct attention mask $\mathbf{\attMSK}_{i,j}$ with  $\mathbf{\tokenstates}_j$, $\mathbf{\EndSeqHardTrue}_j$, and $\SlidingWindowSize$ with Equations~\ref{eq:mskglobal},\ref{eq:msksw},\ref{eq:msklocalfwd}}
        \STATE On chip, compute \textcolor{blue}{$\mathbf{P'}_{i}^{(j)} = \exp(\vS_{ij} - L_{i}) \in \mathbb{R}^{B_r \times B_c}$ and $\vP_{i}^{(j)} = \mathbf{P'}_{i}^{(j)} \odot \mathbf{\attMSK}_{i,j}$.}
        \STATE \textcolor{blue}{Check if the current block contains any valid value}
        \IF {\textcolor{blue}{Do compute}}
        \STATE On chip, compute
        $\vdV_j \leftarrow \vdV_j + (\vP_{i}^{(j)})^\top \vdO_i \in \mathbb{R}^{B_c \times d}$.
        \STATE On chip, compute
        $\vdP_{i}^{(j)} = \vdO_{i} \vV_j^\top \in \mathbb{R}^{B_r \times B_c}$.
        \STATE On chip, compute $\vdS_{i}^{(j)} = \vP_{i}^{(j)} \odot (\vdP_{i}^{(j)} - D_i) \in \mathbb{R}^{B_r \times B_c}$.
        \STATE Load $\vdQ_i$ from HBM to SRAM, then on chip, update
        $\vdQ_{i} \leftarrow \vdQ_i + \vdS_{i}^{(j)} \vK_j \in \mathbb{R}^{B_r \times d}$, and write
        back to HBM.
        \STATE On chip, compute $\vdK_{j} \leftarrow \vdK_j + {\vdS_{i}^{(j)}}^\top \vQ_i \in \mathbb{R}^{B_c \times d}$.
        \ENDIF
        \STATE \textcolor{blue}{On chip, clip $\mathbf{P'}_{i}^{(j)}$ to (0,1) and compute $d\mathbf{S'}_{i}^{(j)} = \mathbf{P'}_{i}^{(j)} \odot (\vdP_{i}^{(j)} - D_i) \in \mathbb{R}^{B_r \times B_c}$}
        \STATE \textcolor{blue}{On chip, update $d\mathbf{\tokenstates}_j$ with Equations~\ref{eq:dmskglobal}, \ref{eq:dmsklocal}, \ref{eq:dmsksw}}
      \ENDFOR
      \STATE Write $\vdK_j, \vdV_j$, \textcolor{blue}{and $d\mathbf{\tokenstates}_j$} to HBM.
    \ENDFOR
    \STATE Return $\vdQ, \vdK, \vdV$.
  \end{algorithmic}
\end{algorithm}

\section{Detailed Inference Process~\label{sec:inferencedetails}}
In Section~\ref{sec:cacheupdate}, we showed that ~\ourname{} could efficiently update the KV cache values and drop the unnecessary tokens. Here we provide a detailed pseudocode for this updating process.

\begin{algorithm}[tbh]
    \caption{\small\label{alg:cacheupdate}\ourname{} KV Cache Updating Process}
    \begin{algorithmic}[1]
    \REQUIRE KV values $\vK, \vV \in \mathbb{R}^{N \times \nheads \times d_{head}}$, sliding window size $\SlidingWindowSize$, $\mathbf{\tokenstates} \in \mathbb{R}^{\seqlen \times \nopts}$. 
    Existing KV cache $\vK^{Cache}, \vV^{Cache} \in \mathbb{R}^{\SlidingWindowSize \times \nheads \times d_{heads}}$, number of global tokens $n^{global} \in \mathbb{R}^{\nheads}$, number of local tokens $n^{local}\in \mathbb{R}^{\nheads}$, current tail of the sliding window queue $t$
    \FOR{$0 \le i \le \seqlen$}
      \FOR {$0 \le h \le \nheads$}
      %\STATE concatenate $\vK_{i}, \vV_{i}$ to the KV caches:$\vK^{Cache} = [\vK^{Cache}, \vK_{i}]$, $\vV^{Cache} = [\vV^{Cache}, \vV_{i}]$
      \STATE remove the tail of the sliding window queue: $\vK^{Cache}_t=0, \vV^{Cache}_t=0$
      \IF{$\tokenstates_{i}$ is $\GlobalToken{}$}
      \STATE move $\vK_{i, h}, \vV_{i, h}$ to the ($n^{global}_h$)th position of the KV cache values: $\vK^{Cache}_{n^{global}_{h}, h} = \vK_{i, h},
      \vV^{Cache}_{n^{global}_{h}, h} = \vV_{i, h},
      $.
      \STATE update number of global and local tokens $n_h^{global} = n_h^{global} + 1$, $n_h^{local} = 0$
      \STATE remove all the KV cache values after $n_h^{global}$: $\vK^{Cache}_{j, h} = 0, \vV^{Cache}_{j, h} = 0, for\ j > n_h^{global}$
      
      \ELSIF{$\tokenstates_{i}$ is $\LocalToken{}$}
      \STATE move $\vK_{i, h}, \vV_{i, h}$ to the ($n^{global}_h + n^{local}_h$)th position of the KV cache values: $\vK^{Cache}_{n^{global}_{h} + n^{local}_{h}, h} = \vK_{i, h},
      \vV^{Cache}_{n^{global}_{h} + n^{local}_{h}, h} = \vV_{i, h},
      $.
      \STATE update number of local tokens $n_h^{local} = n_h^{local} + 1$
      \ELSE
      \STATE move $\vK_{i, h}, \vV_{i, h}$ to the ($t$)th position of the KV cache values: $\vK^{Cache}_{t, h} = \vK_{i, h},
      \vV^{Cache}_{t_{h} + n^{local}_{h}, h} = \vV_{i, h},
      $.
      \ENDIF
      \STATE generate masks for the corresponding valid values
      \STATE move the queue tail to the next value: $t = (t + 1) \bmod \SlidingWindowSize$
      \ENDFOR
    \ENDFOR

    \end{algorithmic}
\end{algorithm}

As shown in Algorithm~\ref {alg:cacheupdate}, once we receive a new KV cache pair, we first check its corresponding type. Depending on the new token type, we either: \begin{enumerate*}
    \item append the new KV values after the next global tokens and remove the remaining local tokens
    \item append the new KV values after the existing tokens
    \item put the new KV values to the tail of the sliding window token queues (located at the beginning of the KV cache values)
\end{enumerate*}

\section{Experiments Details~\label{sec:expdetails}}
Here we discuss further details in our experiments. The codes for ~\ourname{} can be found under ~\url{https://github.com/automl/NeuralAttentionSearch}
\subsection{Detailed Hyperparameter Setting for GPT2-small Training~\label{sec:pg19-train}}
 Following the setting from NanoGPT~\cite{karpathy-github22a}, the GPT-2 style small has 12 layers and 12 heads with a hidden dimension of 768.  Instead of the learnable position encoding, we apply rotary embeddings~\cite{su-nc24a} to each transformer layer. The PG-19 dataset contains books extracted from Project Gutenberg~\cite{gutenberg} with about 2B tokens in the training sets. We train all models with a context length of 1024 and a batch size of 480 (using gradient accumulation). We train them for $600\,000$ iterations evaluate them on the test sets of PG19 with a context length of 1024. Training one model can be finished in 16 hours with 4 Nvidia H100 PCEi GPUs.
 
 %Further details on the hyperparameters can be found in the appendix. 

\subsection{Collecting the Fine-tune Training Set~\label{sec:datacollect}}
To fine-tune ~\ourname{} on LLMs, we collect the training datasets from different tasks for real-world tasks:
\begin{compactitem}
\item Multi-Document QA: HotPotQA~\cite{yang-emnlp18a}, 2WikiMultihopQA~\cite{ho-coling20a}, MuSiQue~\cite{trivedi-tacl22a}, and DuReader (zh)~\cite{he-aclwqa18a}
\item Single-Document QA: NarrativeQA~\cite{kocisky-tacl18a} and Qasper~\cite{dasigi-naacl21a}
\item Summarization: GovReport~\cite{huang-naacl21a}, QMSum~\cite{zhong-naacl21a}, MultiNews~\cite{fabbri-acl19a}, and VCSUM (zh)~\cite{wu-aclf23a}
\item Few-shot Leraning: TREC~\cite{li-coling02a}, TriviaQA~\cite{joshi-acl17a},  and SAMSum~\cite{gliwa-arxiv19a} 
\item Code Completion: LCC~\cite{guo-icml23a} and RepoBench-P~\cite{liu-iclr24b}
\end{compactitem}

We also construct the synthetic passkey-retrieval dataset introduced in DuoAttention~\cite{xiao-arxiv24a}. This dataset is generated by embedding multiple random passkeys in different locations with a long context. The model will then be asked to recall this passkey information.

For all the few-shot learning datasets, following ~\citet{bai-acl24a}, we randomly concatenate multiple question-answer pairs into one single extended context as one piece of data. The number of concatenated samples for the TREC dataset ranges from $[10, 100]$. This value is $[2,6]$ for TriviaQA and $[10, 50]$ for SAMSum. Additionally, for the datasets that do not have enough context length (e.g., the DuReader dataset), we also merge multiple documents into one piece of data (in our case, this value is 4).

We do not collect all the data whose length goes beyond a threshold to ensure that the context can be fitted into our model. Additionally, we collect at most 500 instances in each dataset since some datasets might not contain data. In the end, the real-world dataset contains 6436 data instances. We add another 564 instances from the synthetic dataset. In the end, our dataset contains 7000 instances. Finetuning a model on this dataset for one epoch takes roughly 8 hours with 2 Nvidia H100 PCIe GPUs. 

\section{Further Experimental Results~\label{sec:appres}}

% \subsection{Results on Ruler Benchmark with Mistral}
% \begin{wraptable}[13]{r}{10cm}
% \vspace{-3.5mm}
%     \scalebox{0.65}{\input{tables/ruler/mistral-50}}
%     \scalebox{0.65}{\input{tables/ruler/mistral-25}}
%         \caption{Ruler results with 50\% KV budgets (top) and 25\% KV budget (bottom) on Mistral-7B-Instruct-v0.3. All Full models use the 100\% KV budgets. We mark the actual KV budgets used by ~\ourname{} in the bracket. ~\label{tab:rulermistral}}
% \end{wraptable}

\subsection{Results on Ruler Benchmark with Mistral}
\begin{table}[h]
\centering
    \scalebox{0.8}{\begin{tabular}{l|l|llllllllll}
\toprule
 & Full & Duo & SLLM & H2O & Snap & Ada & Chunk & Critical & Pyradmid & MoA & NAtS \\
\midrule
4k & 93.68 & 92.63 & 58.89 & 5.60 & 46.50 & 57.92 & 59.33 & 74.06 & 44.00 & 55.73 & \textbf {93.68} (49\%) \\
8k & 91.29 & 89.94 & 59.21 & 2.82 & 38.12 & 49.30 & 53.53 & 83.12 & 37.66 & 41.98 & \textbf {90.76} (43\%) \\
16k & 89.85 & 87.51 & 55.18 & 3.00 & 38.59 & 52.80 & 63.46 & 82.83 & 38.05 & 34.31 & \textbf {89.09} (41\%) \\
32k & 81.24 & 78.81 & 47.63 & 2.70 & 49.32 & 66.20 & 71.91 & 76.23 & 46.16 & 27.92 & \textbf {80.39} (42\%) \\
\bottomrule
\end{tabular}
}
    \scalebox{0.8}{\begin{tabular}{l|l|llllllllll}
\toprule
 & Full & Duo & SLLM & H2O & Snap & Ada & Chunk & Critical & Pyradmid & MoA & NAtS \\
\midrule
4k & 93.68 & 52.44 & 42.96 & 3.26 & 33.72 & 38.79 & 42.74 & 46.11 & 33.66 & 33.26 & \textbf {93.18} (29\%) \\
8k & 91.29 & 40.96 & 36.54 & 2.37 & 31.36 & 36.67 & 42.01 & 46.73 & 28.64 & 25.47 & \textbf {89.97} (24\%) \\
16k & 89.85 & 35.42 & 33.38 & 1.92 & 30.30 & 37.48 & 42.84 & 57.17 & 28.17 & 25.55 & \textbf {87.69} (21\%) \\
32k & 81.24 & 40.00 & 32.76 & 1.70 & 35.17 & 44.46 & 60.56 & 61.55 & 31.84 & 24.88 & \textbf {78.60} (21\%) \\
\bottomrule
\end{tabular}
}
        \caption{Ruler results with 50\% KV budgets (top) and 25\% KV budget (bottom) on Mistral-7B-Instruct-v0.3. All Full models use the 100\% KV budgets. We mark the actual KV budgets used by ~\ourname{} in the bracket. ~\label{tab:rulermistral}}
\end{table}

Table~\ref{tab:rulermistral} shows the results of different optimizers on the Mistral-7B-Instruct-v0.3 model. In this case, the $\SparseRegularizedPar$ for 25\% and 50\% budgets are $1e-6$ and $3e-7$ respectively.  We found that to achieve the same sparsity level, Mistral-7B requires a larger sparse regularization value $\SparseRegularizedPar$ compared to the Llama 3.1 model. This might indicate that the attention map values in Mistral are more evenly distributed, and thus we need a larger $\SparseRegularizedPar$ to force more tokens to only focus on the local information. The maximal context length of Mistral 7B is $32k$. We only evaluate the models with a maximal length of $32k$. The result is consistent with the results shown in Table~\ref{tab:rulerllama}. ~\ourname{} consistently outperforms the other baselines with smaller budgets. 

\subsection{Further results on LongBench}

We show additional results on fine-tuning LLM on the LongBench dataset here. Table~\ref{tab:ResLBMistra25} shows the results with 25\% KV budgets on Mistral-7B-Instruct-v0.3 with $\SparseRegularizedPar=3e-6$. The result confirms our conclusion that ~\ourname{} outperforms the other baselines in most datasets under a similar budget level (and many times with even smaller budgets).

\begin{table}[h]
    \centering
    \scalebox{0.75}{\begin{tabular}{l|r|rrrrlrlrrl}
\toprule
 & Full & Duo & SLLM & H2O & Snap & Ada & Chunk & Critical & Pyradmid & MoA & NAtS \\
\midrule
NarrativeQA & 29.21 & 16.49 & 27.56 & 23.65 & 22.70 & 24.34 & 21.22 & \textbf {27.72} & 21.97 & 21.67 & 26.7(9\%) \\
Qasper & 41.34 & 14.58 & 29.16 & 18.95 & 23.08 & 22.99 & 21.21 & 28.85 & 23.54 & 19.72 & \textbf {41.46}(21\%) \\
MultiFieldQA-en & 52.51 & 29.83 & 31.88 & 31.52 & 38.56 & 38.11 & 34.11 & 45.18 & 35.34 & 28.80 & \textbf {52.36}(18\%) \\
MultiFieldQA-zh & 58.03 & 29.15 & 30.47 & 29.21 & 31.42 & 32.30 & 32.40 & 38.12 & 29.67 & 23.20 & \textbf {54.13}(16\%) \\
HotpotQA & 49.62 & 34.85 & 46.07 & 42.10 & 46.37 & 48.83 & 46.93 & 47.48 & 45.94 & 39.64 & \textbf {51.45}(15\%) \\
2WikiQA & 40.01 & 27.83 & 35.11 & 30.43 & 33.22 & 33.14 & 32.03 & \textbf {38.24} & 33.96 & 27.02 & 37.05(18\%) \\
Musique & 28.44 & 13.03 & 22.65 & 16.99 & 22.45 & 21.28 & 22.62 & 25.01 & 23.67 & 16.22 & \textbf {27.88}(14\%) \\
DuReader (zh) & 34.87 & 26.51 & 18.69 & 20.66 & 25.34 & 27.55 & 28.45 & 27.04 & 24.82 & 19.11 & \textbf {35.25}(12\%) \\
GovReport & 34.94 & 22.09 & 27.44 & 27.11 & 28.31 & 27.94 & 28.94 & 29.33 & 26.94 & 25.58 & \textbf {33.39}(14\%) \\
QMSum & 25.67 & 17.25 & 22.11 & 20.84 & 21.43 & 21.95 & 21.95 & 22.77 & 22.39 & 19.53 & \textbf {24.70}(13\%) \\
MultiNews & 27.85 & 23.41 & 22.01 & 23.03 & 23.77 & 23.62 & 23.34 & 24.37 & 23.89 & 22.55 & \textbf {27.35}(28\%) \\
VCSUM (zh) & 16.34 & 13.58 & 15.41 & 15.13 & 15.28 & 15.38 & 14.69 & \textbf {16.10} & 14.96 & 14.87 & 16.06(11\%) \\
TREC & 75.50 & 52.00 & 70.50 & 61.50 & 58.50 & 59.00 & 57.00 & 67.50 & 58.00 & 63.00 & \textbf {73.50}(20\%) \\
TriviaQA & 88.89 & 84.85 & 88.92 & 86.60 & 88.89 & 89.39 & 89.19 & 88.98 & 89.31 & 86.94 & \textbf {89.80}(17\%) \\
SAMSum & 47.29 & 42.15 & 45.44 & 45.16 & 46.40 & 46.72 & 46.51 & 46.67 & 46.58 & 43.90 & \textbf {46.91}(13\%) \\
LSHT & 39.75 & 17.50 & 29.00 & 22.00 & 38.50 & 39.00 & \textbf {39.50} & 39.25 & 37.50 & 17.25 & 38.5(11\%) \\
Passage Count & 5.50 & 3.50 & 6.00 & 6.00 & 6.00 & \textbf {8.00} & 5.50 & 6.00 & 6.50 & 5.00 & 4.5(14\%) \\
PassageRetrieval-en & 98.00 & 62.50 & 86.50 & 63.00 & 91.50 & 94.00 & 87.50 & 89.50 & 88.00 & 17.25 & \textbf {96.00}(13\%) \\
PassageRetrieval-zh & 96.50 & 10.50 & 26.50 & 34.50 & 69.25 & 83.25 & 71.00 & 90.00 & 75.50 & 16.00 & \textbf {92.00}(14\%) \\
LCC & 53.02 & 50.87 & 51.22 & 52.77 & 55.67 & 55.67 & 53.76 & \textbf {56.17} & 55.46 & 51.08 & 53.81(29\%) \\
RepoBench-P & 56.83 & 48.93 & 54.80 & 55.74 & 56.82 & \textbf {57.06} & 56.75 & 56.75 & 56.40 & 53.59 & 56.7(23\%) \\
\bottomrule
\end{tabular}
}
    \caption{LongBench Results with 25\% Budget Size on Mistral-7B.~\label{tab:ResLBMistra25}.}
\end{table}

\begin{table}
    \centering
    \scalebox{0.75}{\begin{tabular}{l|r|lrllrrlrrl}
\toprule
 & Full & Duo & SLLM & H2O & Snap & Ada & Chunk & Critical & Pyradmid & MoA & NAtS \\
\midrule
NarrativeQA & 31.35 & 29.40 & 28.18 & 20.15 & 30.37 & 30.04 & 29.49 & \textbf {32.03} & 29.18 & 13.92 & 31.39(29\%) \\
Qasper & 24.73 & 20.37 & 18.15 & 18.09 & 18.44 & 19.42 & 17.24 & 22.13 & 18.54 & 22.25 & \textbf {25.17}(45\%) \\
MultiFieldQA-en & 29.46 & 26.70 & 17.89 & 22.50 & 23.21 & 24.62 & 22.55 & 26.30 & 22.95 & 14.24 & \textbf {28.94}(41\%) \\
MultiFieldQA-zh & 60.01 & 60.31 & 42.19 & 38.88 & 47.66 & 50.23 & 44.54 & 56.22 & 46.50 & 23.91 & \textbf {61.27}(37\%) \\
HotpotQA & 17.06 & \textbf {18.84} & 16.73 & 15.61 & 16.55 & 16.29 & 17.22 & 15.83 & 15.86 & 8.85 & 17.28(37\%) \\
2WikiQA & 16.64 & 16.30 & 14.28 & 11.71 & 15.82 & 15.38 & 14.67 & \textbf {17.01} & 14.81 & 10.58 & 16.84(41\%) \\
Musique & 11.59 & \textbf {13.77} & 10.75 & 10.12 & 10.78 & 11.19 & 10.63 & 12.85 & 10.64 & 5.03 & 11.89(36\%) \\
DuReader (zh) & 35.56 & 33.29 & 18.16 & 27.32 & 29.22 & 30.82 & 28.49 & 32.81 & 27.93 & 22.80 & \textbf {33.36}(32\%) \\
GovReport & 34.30 & 32.99 & 30.84 & 29.46 & 31.34 & 31.51 & 31.92 & 33.09 & 30.54 & 25.45 & \textbf {34.23}(36\%) \\
QMSum & 23.30 & \textbf {23.89} & 21.43 & 20.72 & 22.08 & 22.74 & 22.22 & 22.89 & 22.92 & 19.76 & 22.6(33\%) \\
MultiNews & 27.13 & 26.29 & 24.77 & 25.10 & 25.27 & 25.48 & 25.08 & 25.81 & 25.10 & 24.86 & \textbf {27.17}(51\%) \\
VCSUM (zh) & 16.36 & 15.70 & 15.29 & 15.03 & 15.70 & 15.77 & 15.59 & 15.89 & 15.32 & 14.67 & \textbf {16.24}(28\%) \\
TREC & 72.50 & 72.50 & 71.00 & 65.50 & 60.00 & 66.50 & 63.50 & 70.50 & 59.50 & 64.00 & \textbf {73.50}(44\%) \\
TriviaQA & 91.15 & 90.41 & 91.44 & 90.25 & 91.47 & 90.97 & 91.63 & 91.14 & 90.98 & 81.17 & \textbf {92.14}(40\%) \\
SAMSum & 43.72 & 42.68 & 43.41 & 42.57 & \textbf {44.21} & 43.44 & 43.00 & 43.41 & 43.91 & 39.64 & 42.61(30\%) \\
LSHT & 46.50 & \textbf {46.50} & 40.00 & 34.50 & 45.00 & 46.00 & 45.00 & 46.50 & 45.00 & 24.00 & 45.5(31\%) \\
Passage Count & 6.63 & 6.67 & 6.73 & 2.21 & 6.33 & 6.07 & 6.03 & 4.74 & 8.04 & 1.80 & \textbf {9.78}(37\%) \\
PassageRetrieval-en & 97.98 & \textbf {98.55} & 96.67 & 93.54 & 95.98 & 97.88 & 95.83 & 98.12 & 96.10 & 17.96 & 97.32(36\%) \\
PassageRetrieval-zh & 77.99 & 75.58 & 43.61 & 67.83 & \textbf {79.66} & 78.27 & 72.38 & 78.22 & 79.48 & 19.72 & 78.17(34\%) \\
LCC & 54.10 & \textbf {56.22} & 53.10 & 55.61 & 54.08 & 53.13 & 54.23 & 52.33 & 51.87 & 52.28 & 53.27(57\%) \\
RepoBench-P & 51.39 & \textbf {57.54} & 50.80 & 55.17 & 52.49 & 51.59 & 52.32 & 52.38 & 51.59 & 48.46 & 52.68(47\%) \\
\bottomrule
\end{tabular}
}
    \caption{LongBench Results with 50\% Budget Size on LLama 3.1 8B~\label{tab:ResLBMLLama50}.}
\end{table}

\begin{table}
    \centering
    \scalebox{0.75}{\begin{tabular}{l|r|lllrrlllrl}
\toprule
 & Full & Duo & SLLM & H2O & Snap & Ada & Chunk & Critical & Pyradmid & MoA & NAtS \\
\midrule
NarrativeQA & 29.21 & 26.95 & \textbf {28.76} & 24.00 & 24.65 & 25.40 & 25.66 & 28.20 & 24.41 & 23.58 & 28.1(31\%) \\
Qasper & 41.34 & 35.65 & 35.80 & 27.88 & 31.68 & 31.94 & 31.38 & 38.10 & 31.50 & 28.60 & \textbf {41.75}(46\%) \\
MultiFieldQA-en & 52.51 & 52.50 & 37.35 & 40.15 & 46.12 & 48.52 & 44.73 & 52.20 & 45.70 & 35.52 & \textbf {52.79}(42\%) \\
MultiFieldQA-zh & 58.03 & 55.37 & 36.43 & 39.43 & 40.55 & 40.92 & 43.70 & 47.93 & 38.24 & 27.48 & \textbf {56.75}(38\%) \\
HotpotQA & 49.62 & \textbf {52.94} & 47.66 & 47.24 & 49.34 & 49.57 & 49.51 & 47.20 & 51.26 & 39.64 & 50.73(39\%) \\
2WikiQA & 40.01 & 39.26 & 38.44 & 38.09 & 37.27 & 36.75 & 40.24 & 38.13 & 37.49 & 31.42 & \textbf {40.77}(42\%) \\
Musique & 28.44 & \textbf {29.45} & 27.24 & 20.61 & 26.43 & 25.77 & 27.28 & 28.54 & 25.85 & 16.84 & 28.64(38\%) \\
DuReader (zh) & 34.87 & \textbf {36.44} & 18.63 & 26.82 & 30.29 & 32.18 & 32.72 & 32.18 & 29.71 & 20.79 & 34.7(33\%) \\
GovReport & 34.94 & 32.07 & 30.79 & 30.58 & 31.43 & 31.55 & 32.16 & 32.39 & 29.63 & 25.80 & \textbf {34.23}(38\%) \\
QMSum & 25.67 & 24.20 & 22.91 & 22.48 & 23.16 & 24.36 & 23.47 & 25.12 & 24.11 & 21.08 & \textbf {25.94}(37\%) \\
MultiNews & 27.85 & 26.97 & 25.18 & 25.62 & 25.97 & 26.24 & 25.83 & 26.26 & 25.56 & 24.39 & \textbf {27.47}(54\%) \\
VCSUM (zh) & 16.34 & 15.58 & 16.24 & 15.80 & 15.94 & 16.42 & 15.73 & \textbf {16.57} & 15.63 & 14.13 & 16.55(31\%) \\
TREC & 75.50 & 74.50 & 74.00 & 66.00 & 66.50 & 68.50 & 69.00 & 74.50 & 64.50 & 69.00 & \textbf {75.00}(49\%) \\
TriviaQA & 88.89 & 87.37 & 89.06 & 88.36 & 88.71 & 88.81 & 88.69 & 88.08 & \textbf {89.55} & 87.05 & 89.3(41\%) \\
SAMSum & 47.29 & 44.85 & 47.30 & 45.82 & 47.19 & 47.30 & 46.77 & \textbf {47.31} & 46.85 & 45.29 & 46.87(35\%) \\
LSHT & 39.75 & 37.50 & 33.00 & 31.50 & 39.25 & 39.25 & 39.25 & \textbf {40.25} & 38.00 & 21.00 & 38.75(32\%) \\
Passage Count & 5.50 & 6.00 & 5.50 & 4.00 & 4.00 & 5.50 & 5.00 & 5.00 & 4.00 & 4.50 & \textbf {7.50}(39\%) \\
PassageRetrieval-en & 98.00 & \textbf {99.00} & 89.50 & 81.00 & 97.50 & 98.50 & 97.00 & 98.00 & 98.50 & 33.00 & 98.0(35\%) \\
PassageRetrieval-zh & 96.50 & 96.50 & 51.50 & 76.00 & 89.75 & 95.50 & 92.00 & \textbf {97.00} & 96.50 & 17.50 & 96.0(35\%) \\
LCC & 53.02 & 53.22 & 52.94 & 54.40 & 54.50 & 54.14 & 53.79 & \textbf {54.63} & 53.78 & 52.04 & 53.66(56\%) \\
RepoBench-P & 56.83 & 55.89 & 55.47 & \textbf {57.55} & 56.76 & 57.05 & 55.68 & 56.38 & 56.36 & 54.67 & 57.38(52\%) \\
\bottomrule
\end{tabular}}
    \caption{LongBench Results with 50\% Budget Size on Mistral-7B.~\label{tab:ResLBMistra50}.}
\end{table}

Tables ~\ref{tab:ResLBMLLama50} and ~\ref{tab:ResLBMistra50} show the evaluation results on LongBench with 50\% budgets. Despite having fewer KV cache budgets in all the datasets, ~\ourname{} still achieves comparable performance on the LLama3-8B model and better results on the Mistral model and generally performs comparable to the results with the full attention transformers.

\section{Ablation Study}

\subsection{Sparse Regularization Term ~$\SparseRegularizedPar$}
\begin{table}
    \centering
    \scalebox{0.8}{\begin{tabular}{l|r|lllll}
\toprule
 & Full & NAtS 1e-6 & NAtS 5e-7 & NAtS 3e-7 & NAtS 1e-7 & NAtS 5e-8 \\
\midrule
NarrativeQA & 31.4 & 27.05(5\%) & 29.19(9\%) & \underline{30.53}(13\%) & \underline{31.39}(29\%) & \underline{31.94}(41\%) \\
Qasper & 24.7 & \underline{\textbf{31.39}}(15\%) & \underline{\textbf{28.6}}(21\%) & \underline{\textbf{23.76}}(27\%) & \underline{\textbf{25.17}}(45\%) & \underline{\textbf{26.97}}(58\%) \\
MultiFieldQA-en & 29.5 & \underline{\textbf{26.84}}(13\%) & \underline{\textbf{27.41}}(18\%) & \underline{\textbf{28.94}}(24\%) & \underline{\textbf{28.94}}(41\%) & \underline{\textbf{29.7}}(54\%) \\
MultiFieldQA-zh & 60.0 & 51.73(14\%) & \underline{56.64}(17\%) & \underline{\textbf{61.18}}(22\%) & \underline{\textbf{61.27}}(37\%) & \underline{\textbf{61.87}}(49\%) \\
HotpotQA & 17.1 & 15.47(10\%) & \underline{17.44}(15\%) & 16.06(20\%) & \underline{17.28}(37\%) & 16.65(49\%) \\
2WikiQA & 16.6 & 15.37(13\%) & \underline{16.44}(18\%) & \underline{\textbf{17.05}}(23\%) & \underline{16.84}(41\%) & \underline{16.75}(53\%) \\
Musique & 11.6 & 10.78(9\%) & 9.87(13\%) & \underline{11.42}(18\%) & \underline{11.89}(36\%) & \underline{11.31}(48\%) \\
DuReader (zh) & 35.6 & 30.06(9\%) & \underline{\textbf{33.63}}(13\%) & \underline{\textbf{34.28}}(17\%) & \underline{\textbf{33.36}}(32\%) & \underline{\textbf{34.41}}(45\%) \\
GovReport & 34.3 & \underline{30.82}(10\%) & \underline{32.47}(14\%) & \underline{\textbf{33.82}}(19\%) & \underline{\textbf{34.23}}(36\%) & \underline{\textbf{34.8}}(50\%) \\
QMSum & 23.3 & \underline{23.37}(8\%) & \underline{23.09}(12\%) & \underline{23.11}(17\%) & \underline{22.6}(33\%) & \underline{22.74}(45\%) \\
MultiNews & 27.1 & \underline{\textbf{26.45}}(22\%) & \underline{\textbf{26.57}}(28\%) & \underline{\textbf{26.72}}(33\%) & \underline{\textbf{27.17}}(51\%) & \underline{\textbf{27.11}}(64\%) \\
VCSUM (zh) & 16.4 & \underline{15.56}(7\%) & \underline{15.61}(10\%) & \underline{15.61}(14\%) & \underline{\textbf{16.24}}(28\%) & \underline{\textbf{15.99}}(39\%) \\
TREC & 72.5 & \underline{68.0}(14\%) & \underline{70.5}(20\%) & \underline{72.0}(26\%) & \underline{\textbf{73.5}}(44\%) & \underline{\textbf{73.0}}(56\%) \\
TriviaQA & 91.2 & 91.38(12\%) & 90.32(17\%) & \underline{91.61}(22\%) & \underline{\textbf{92.14}}(40\%) & \underline{\textbf{91.64}}(52\%) \\
SAMSum & 43.7 & \underline{43.98}(8\%) & \underline{44.0}(12\%) & \underline{44.21}(16\%) & 42.61(30\%) & \underline{43.84}(41\%) \\
LSHT & 46.5 & 46.0(8\%) & 44.0(12\%) & \underline{\textbf{47.5}}(16\%) & 45.5(31\%) & 45.5(43\%) \\
Passage Count & 6.6 & 2.56(10\%) & 6.88(14\%) & 7.12(20\%) & \underline{\textbf{9.78}}(37\%) & \underline{\textbf{8.82}}(50\%) \\
PassageRetrieval-en & 98.0 & 48.17(9\%) & 82.65(14\%) & \underline{95.81}(19\%) & \underline{97.32}(36\%) & \underline{96.93}(48\%) \\
PassageRetrieval-zh & 78.0 & 33.14(12\%) & 64.77(16\%) & 78.5(20\%) & 78.17(34\%) & 77.7(46\%) \\
LCC & 54.1 & \underline{55.46}(24\%) & 54.53(31\%) & 53.54(38\%) & 53.27(57\%) & 54.4(69\%) \\
RepoBench-P & 51.4 & 50.78(14\%) & 51.85(21\%) & 53.48(28\%) & 52.68(47\%) & 52.36(59\%) \\
\bottomrule
\end{tabular}
} 
    \caption{Ablation Study of Sparse Regularization values $\SparseRegularizedPar$ for LLama3.1-8B~\label{tab:AblationlambdaLLama3}}
\end{table}

% \begin{table}[H]
%    \centering
%     \scalebox{0.75}{\input{tables/ablation/alpha/mistral-lambda}} 
%     \caption{Ablation Study of Sparse Regularization values $\SparseRegularizedPar$ for Mistal-7B~\label{tab:AblationlambdaMistral}}
% \end{table}

% We first study the impact of sparse regularization terms. The result is shown in Table ~\ref{tab:AblationlambdaLLama3} and ~\ref{tab:AblationlambdaMistral}. We underline the results that are better than the optimal baselines with $25\%$ budgets, and bold the results that are better than the optimal baselines with $50\%$. Despite that \ourname{}  in Table~\ref{tab:ResLLBlama25} ($\ourname{}\  1e-7$) and Table~\ref{tab:ResLBMistra25} ($\ourname{}\  1e-6$) used more than $25\%$ overall KV budgets for some tasks, here we show $\ourname{}$ could still outperform many of the corresponding optimal baselines with a even lower KV budget. 

% Table ~\ref{tab:AblationlambdaLLama3} and ~\ref{tab:AblationlambdaMistral} show that stronger $\SparseRegularizedPar$ generally results in a smaller valid KV cache size. While the order for compression rates for different datasets is consistent with different $\SparseRegularizedPar$ settings.  For most tasks, a compression rate between $10\%$ to $20\%$ already results in predictions that are similar to the full attention. 

We first study the impact of sparse regularization terms $\SparseRegularizedPar$. This value controls the efficient KV values cached in the model and thus the model performance. The result is shown in Table~\ref{tab:AblationlambdaLLama3}.  We underline the results that are better than the optimal baselines with $25\%$ budgets, and bold the results that are better than the optimal baselines with $50\%$. Despite that \ourname{}  in Table~\ref{tab:ResLLBlama25} ($\ourname{}\  3e-7$) used more than $25\%$ overall KV budgets for some tasks, here we show $\ourname{}$ could still outperform many of the corresponding optimal baselines with an even smaller KV budget.

\subsection{Sliding Window Length \SlidingWindowSize}
Another important hyperparameter for \ourname{} is the sliding window size  \SlidingWindowSize.  We apply different sliding window sizes \SlidingWindowSize{} ($64, 128, 256, 512$) to fine-tune the Llama3-8B model (with  $\SparseRegularizedPar=3e-7$).

\begin{table}[h]
    \centering
    \scalebox{0.8}{\begin{tabular}{l|r|llll}
\toprule
 & Full & NAtS 64 & NAtS 128 & NAtS 256 & NAtS 512 \\
\midrule
NarrativeQA & 31.4 & 21.61(33\%) & 30.76(15\%) & 30.53(13\%) & 28.96(12\%) \\
Qasper & 24.7 & 17.92(34\%) & 24.1(28\%) & 23.76(27\%) & 28.24(29\%) \\
MultiFieldQA-en & 29.5 & 19.37(34\%) & 29.02(26\%) & 28.94(24\%) & 25.87(25\%) \\
MultiFieldQA-zh & 60.0 & 35.49(35\%) & 60.37(22\%) & 61.18(22\%) & 58.08(26\%) \\
HotpotQA & 17.1 & 13.99(33\%) & 16.41(23\%) & 16.06(20\%) & 17.44(18\%) \\
2WikiQA & 16.6 & 11.52(34\%) & 16.69(26\%) & 17.05(23\%) & 16.48(24\%) \\
Musique & 11.6 & 9.42(33\%) & 11.62(22\%) & 11.42(18\%) & 10.89(16\%) \\
DuReader (zh) & 35.6 & 22.26(34\%) & 35.75(19\%) & 34.28(17\%) & 33.91(16\%) \\
GovReport & 34.3 & 26.63(34\%) & 33.89(22\%) & 33.82(19\%) & 33.05(19\%) \\
QMSum & 23.3 & 20.46(33\%) & 23.52(19\%) & 23.11(17\%) & 22.66(16\%) \\
MultiNews & 27.1 & 23.78(36\%) & 26.73(33\%) & 26.72(33\%) & 26.46(40\%) \\
VCSUM (zh) & 16.4 & 13.8(34\%) & 16.1(15\%) & 15.61(14\%) & 15.9(15\%) \\
TREC & 72.5 & 46.5(34\%) & 73.0(31\%) & 72.0(26\%) & 70.5(25\%) \\
TriviaQA & 91.2 & 84.44(34\%) & 91.89(26\%) & 91.61(22\%) & 91.66(21\%) \\
SAMSum & 43.7 & 38.97(34\%) & 43.62(18\%) & 44.21(16\%) & 44.47(17\%) \\
LSHT & 46.5 & 26.5(34\%) & 47.0(18\%) & 47.5(16\%) & 45.5(15\%) \\
Passage Count & 6.6 & 0.0(34\%) & 5.75(23\%) & 7.12(20\%) & 5.66(18\%) \\
PassageRetrieval-en & 98.0 & 7.13(33\%) & 96.71(23\%) & 95.81(19\%) & 85.78(17\%) \\
PassageRetrieval-zh & 78.0 & 5.15(34\%) & 81.64(21\%) & 78.5(20\%) & 70.66(21\%) \\
LCC & 54.1 & 39.13(36\%) & 54.92(39\%) & 53.54(38\%) & 54.15(44\%) \\
RepoBench-P & 51.4 & 36.87(34\%) & 51.89(32\%) & 53.48(28\%) & 51.34(26\%) \\
\bottomrule
\end{tabular}
}
        \caption{Ablation Study of Sliding Window Length $\SlidingWindowSize$ for Llama3~\label{tab:AblationSWLLama}}
\end{table}

 The result is shown in Table~\ref{tab:AblationSWLLama}. Overall, all the approaches perform similarly. However, a smaller sliding window size generally results in an overall larger KV cache size. A reduced sliding window size would force the model to apply more \GlobalTokens{} and \LocalTokens{} to construct the mid-range correlation since this distance cannot be covered by the sliding window tokens. However, as the number of sliding window sizes further increases, this compression rate might saturate since the remaining tokens might always require a long-range correlation whose distance is much larger than the sliding window size (e.g., these tokens might require the correlation between two tokens whose distances are larger than 1k or even more).

\section{KV Size Distributions}

\begin{figure}[h]
    \centering
    \centering
    \subfigure{\includegraphics[width=0.48\textwidth]{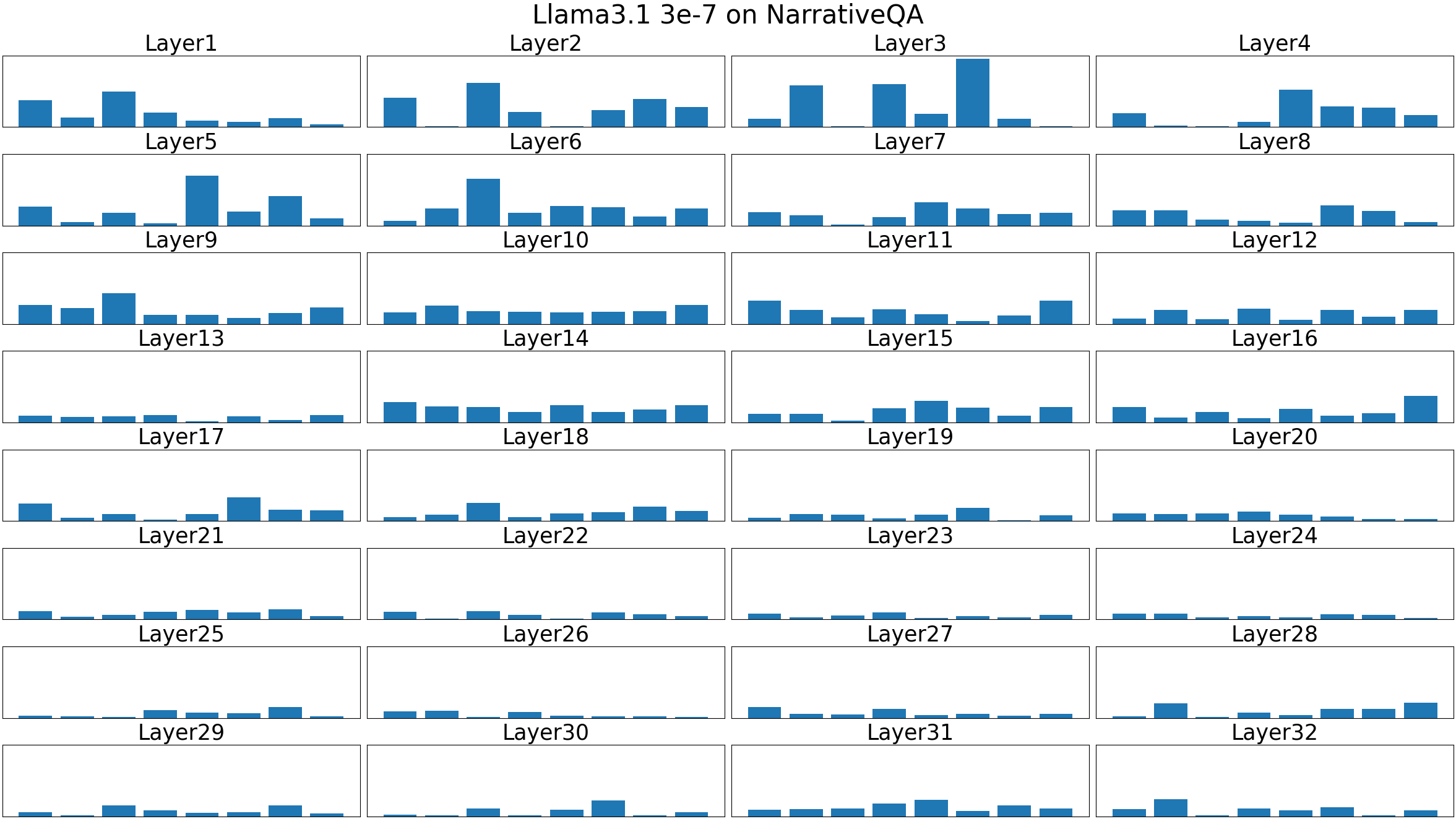}}
    \hfill
    \subfigure{\includegraphics[width=0.48\textwidth]{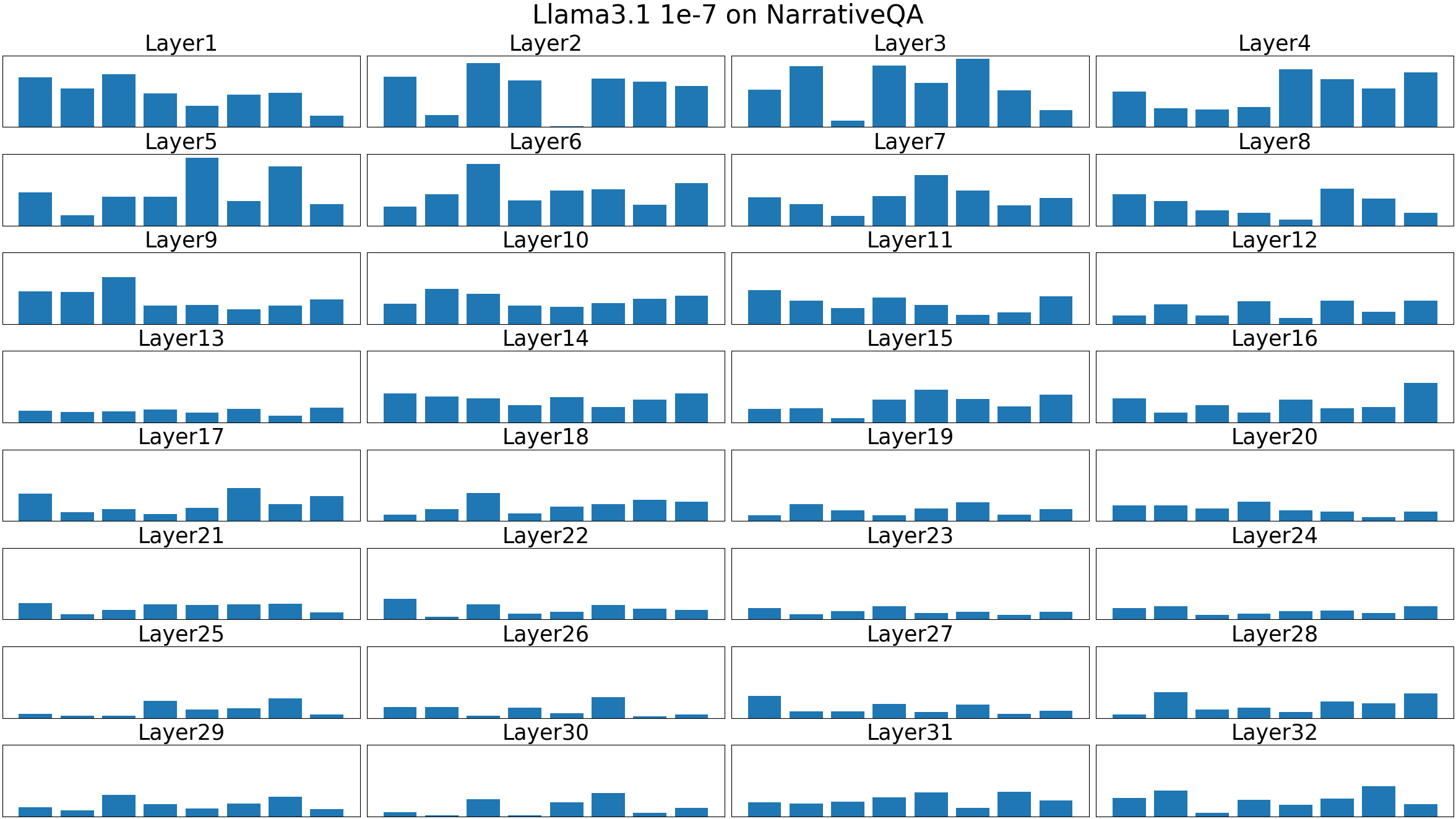}}
    \hfill
    \caption{KV size of LLama on Narrative QA Dataset~\label{fig:kvsizeNarrativeqaLLama}}
\end{figure}

\begin{figure}[h]
    \centering
    \subfigure{\includegraphics[width=0.48\textwidth]{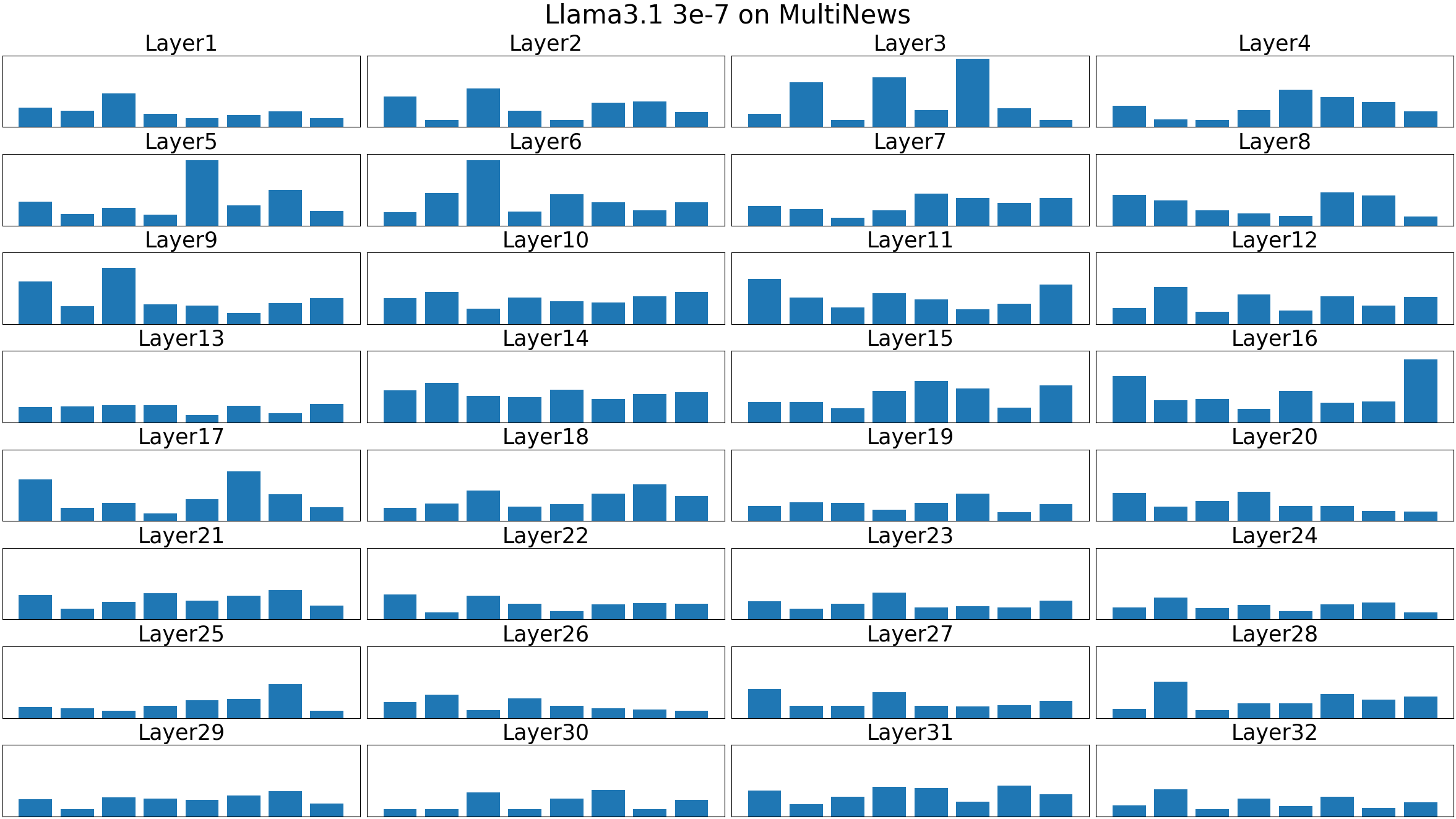}}
    \hfill
    \subfigure{\includegraphics[width=0.48\textwidth]{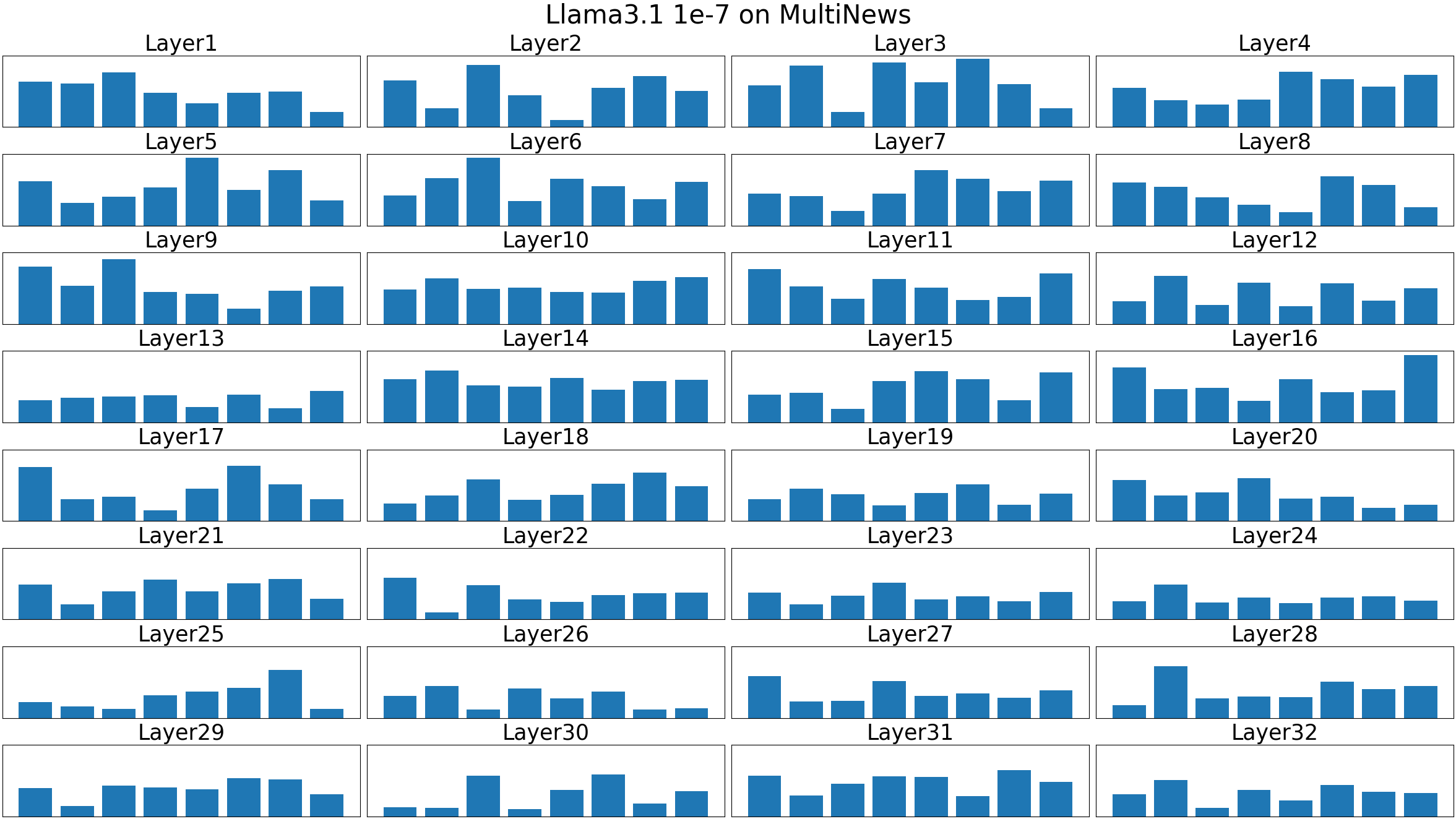}}
    \hfill
    \caption{KV size of LLama on Multi-News Dataset~\label{fig:kvsizeMultiNewsLLama}}
\end{figure}

Here, we provide more KV distribution results with different datasets and hyperparameters on the longbench dataset. 

Figures~\ref{fig:kvsizeNarrativeqaLLama} and ~\ref{fig:kvsizeMultiNewsLLama} show the KV cache size distributions for Llama 3.1 8B on the Narrative QA and Multi News Datasets. The models tend to preserve more KV caches in the shallower layers and ignore the KV caches in the intermediate layers. However, the model still preserves several KV caches near the output layers. Additionally, even in the same layer, the distributions of the KV cache sizes are not evenly distributed in each layer. Some heads are always preferred while the others might be dropped as the context or $\SparseRegularizedPar$ changes.

\begin{figure}[H]
    \centering
    \subfigure{\includegraphics[width=0.48\textwidth]{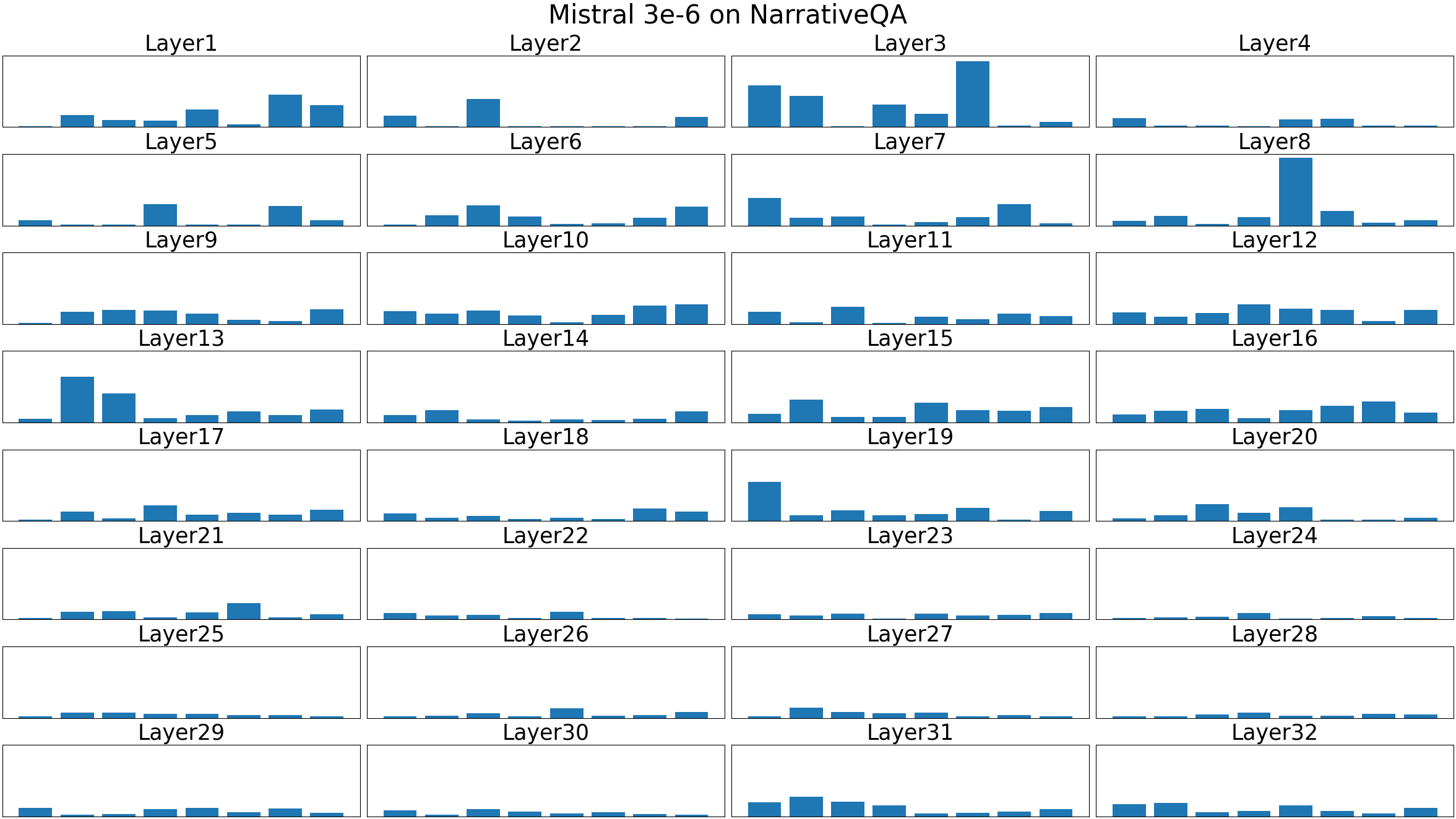}}
    \hfill
    \subfigure{\includegraphics[width=0.48\textwidth]{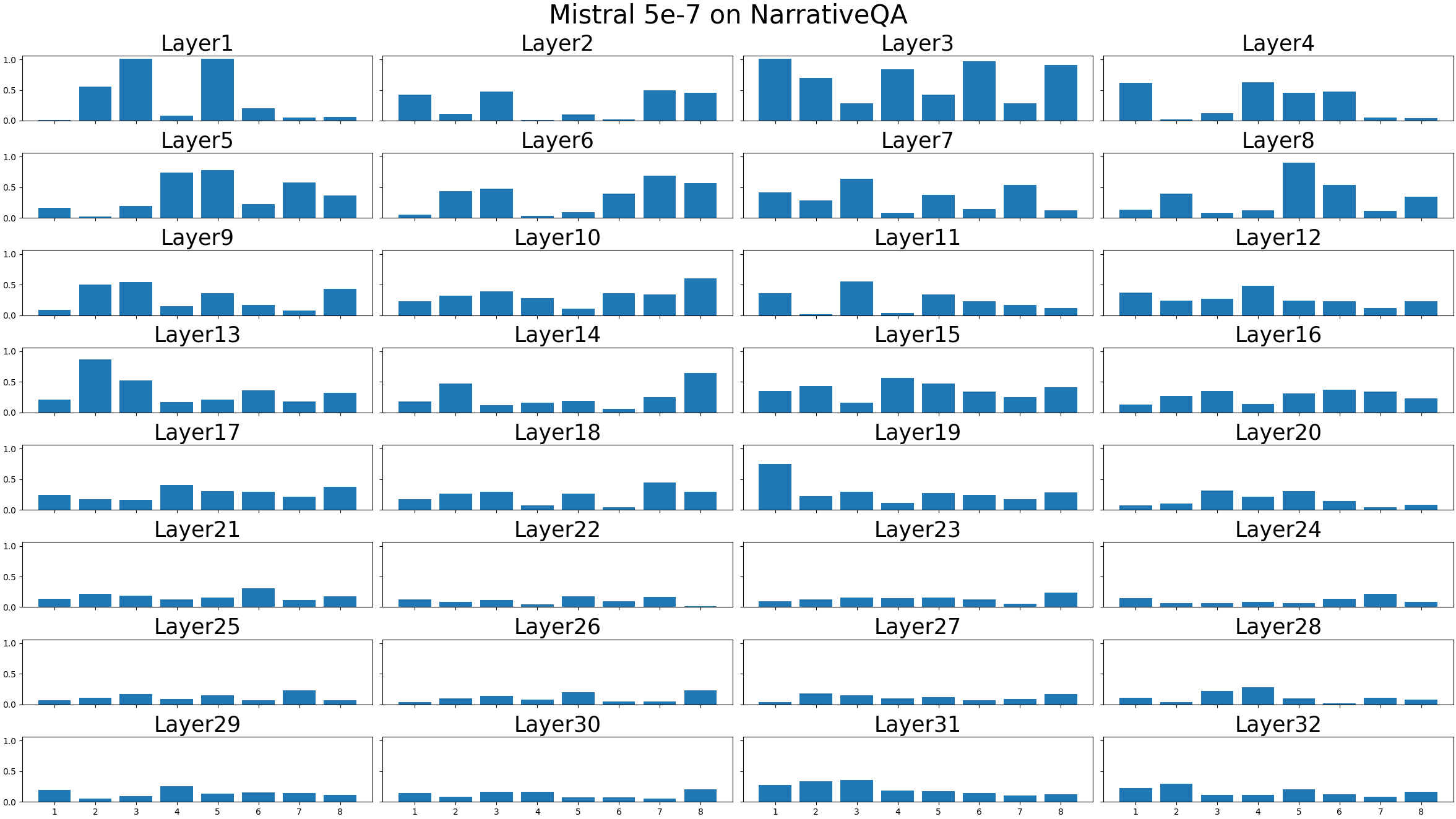}}
    \hfill
    \caption{KV size of Mistral on NarrativeQA Dataset~\label{fig:kvsizeNarrativeqaMistral}}
\end{figure}

\begin{figure}[H]
    \centering
    \subfigure{\includegraphics[width=0.48\textwidth]{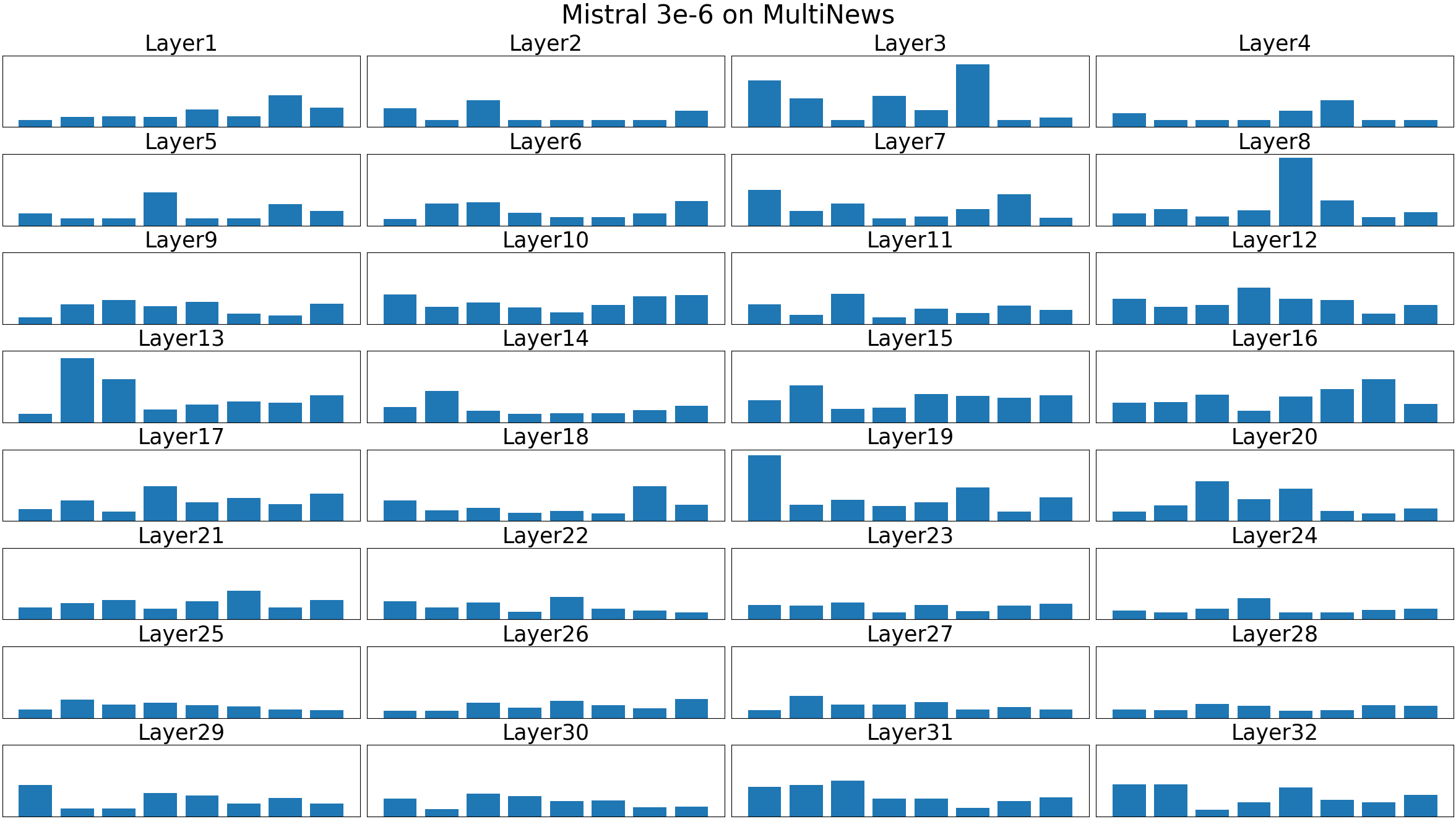}}
    \hfill
    \subfigure{\includegraphics[width=0.48\textwidth]{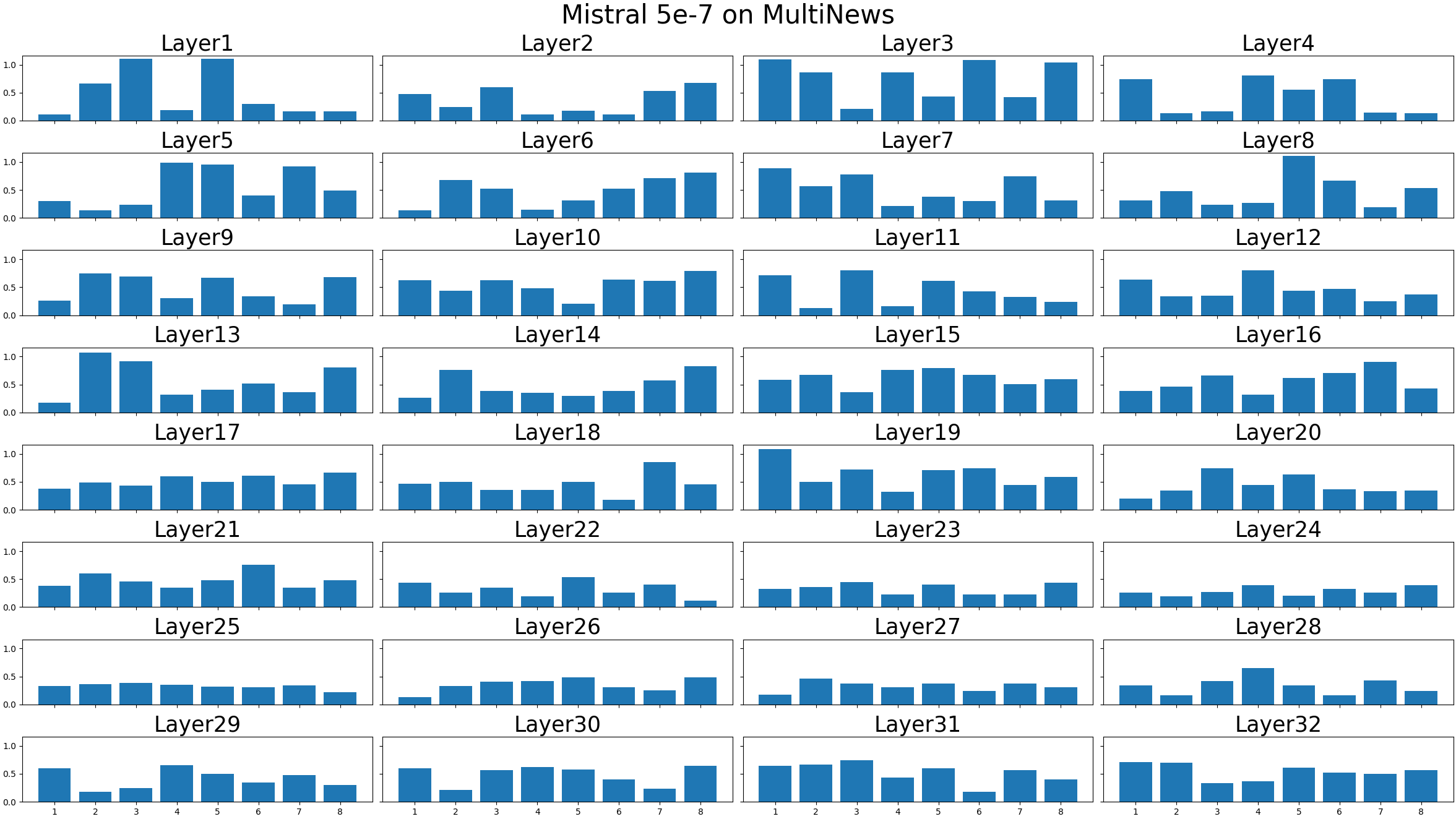}}
    \hfill
    \caption{KV size of Mistral on Multi-News Dataset~\label{fig:kvsizemultinewsmistral}}
\end{figure}

Figures ~\ref{fig:kvsizeNarrativeqaMistral} and ~\ref{fig:kvsizemultinewsmistral} show the KV cache size distributions from the Mistral model. Although the two models share some similar trends, e.g., both models tend to preserve more tokens in the shallower and the last few layers. However, compared to Llama, the KV caches for the Mistral model are more unevenly distributed and tend to gather towards some specific heads. Despite that, both Mistral models and LLama models are GQA models~\citep{ainslie-emnlp23a} and share similar architectures, the preserved token distributions can still be different. This shows that the KV importance distributions might not only depend on the architectures, but are also closely related to the model weights. This highlights the importance of adapting different KV eviction strategies to different models. 

\begin{figure}[H]
    \centering
    \subfigure{\includegraphics[width=0.48\textwidth]{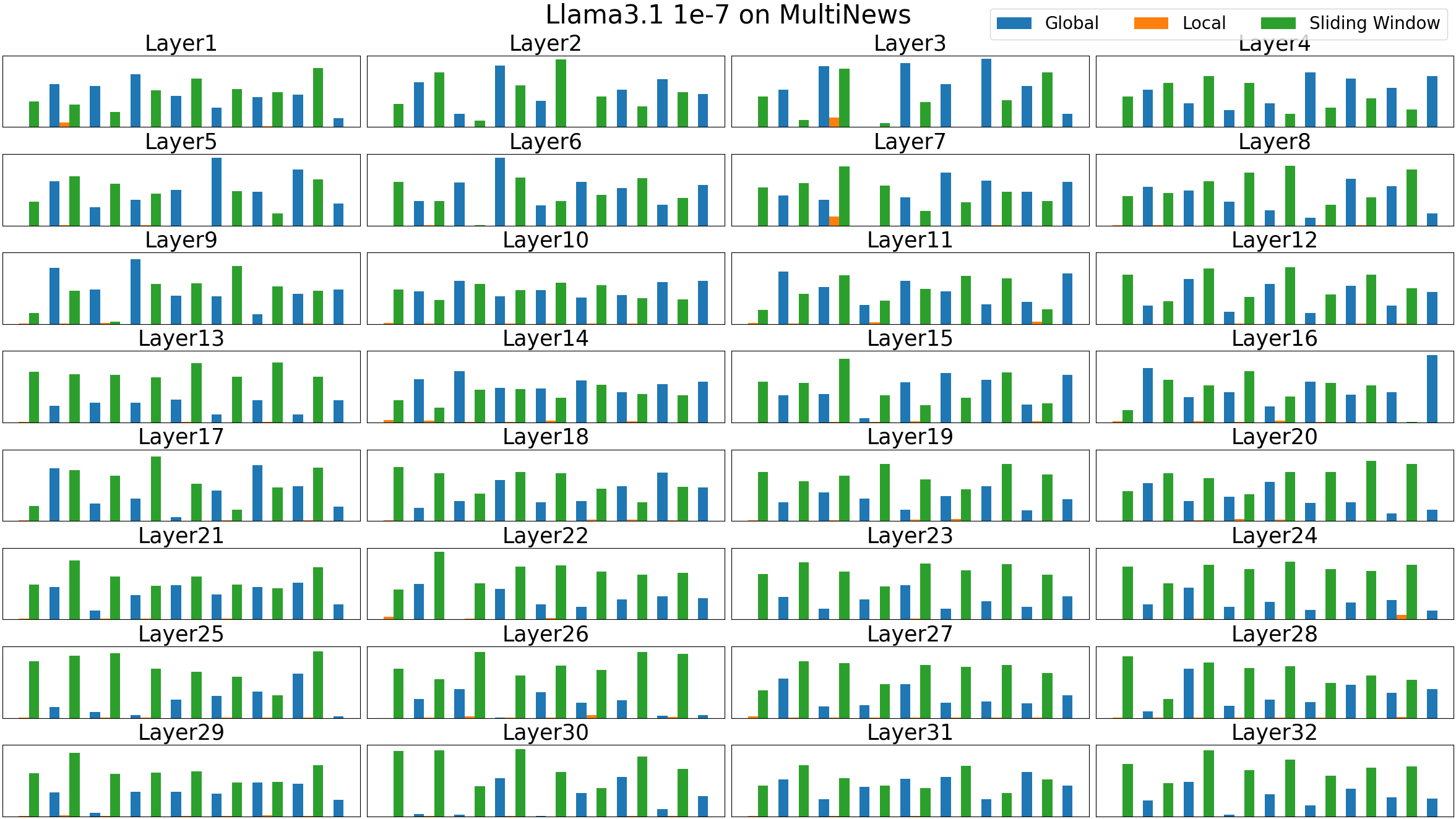}}
    \hfill
    \subfigure{\includegraphics[width=0.48\textwidth]{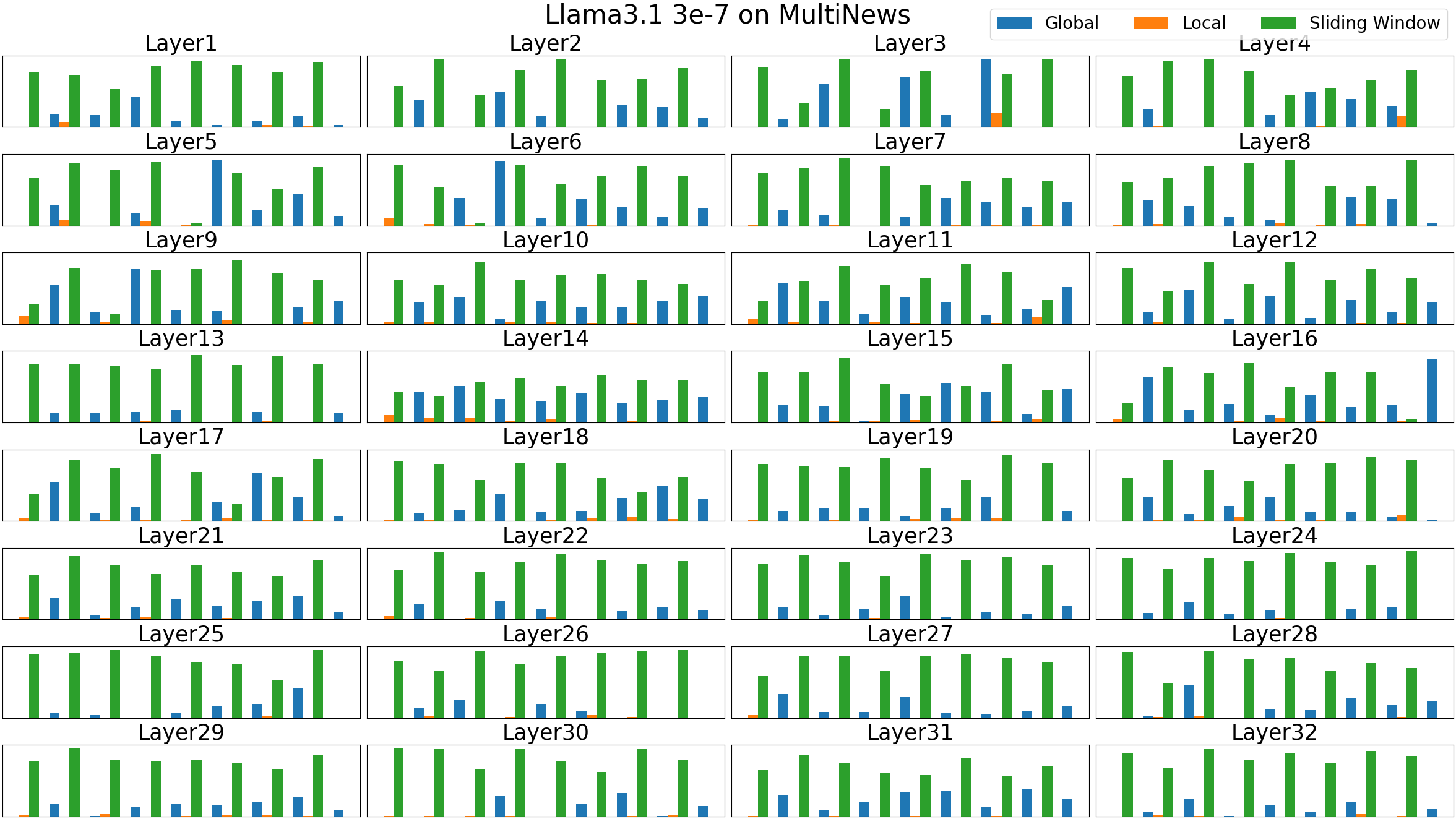}}
    \hfill
    \caption{Number of Token Types for Multi-News Dataset~\label{fig:ntokentypesmultinew}}
\end{figure}

\section{Token Roles Distributions}
\begin{figure}[H]
    \centering
    \subfigure{\includegraphics[width=0.48\textwidth]{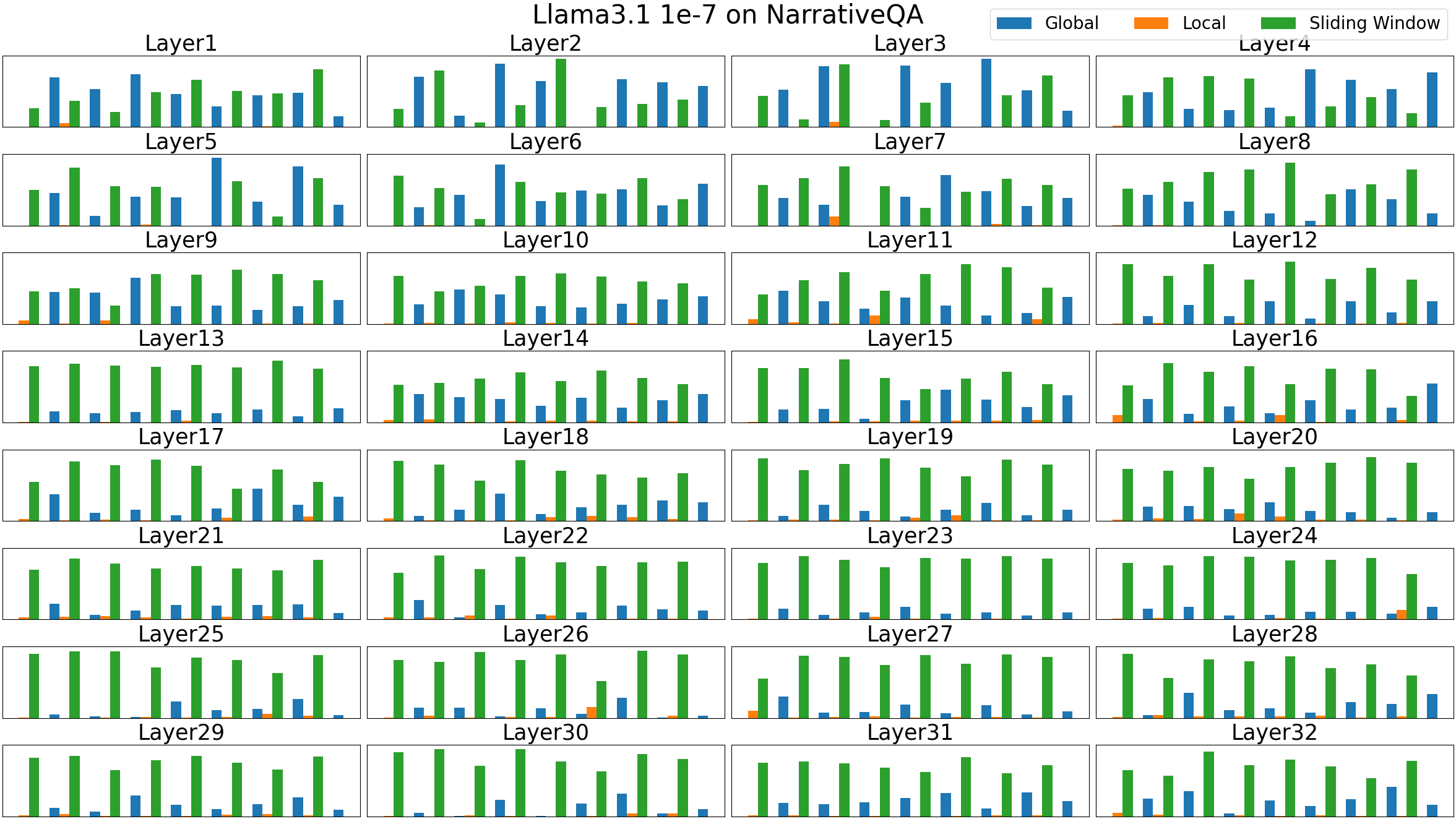}}
    \hfill
    \subfigure{\includegraphics[width=0.48\textwidth]{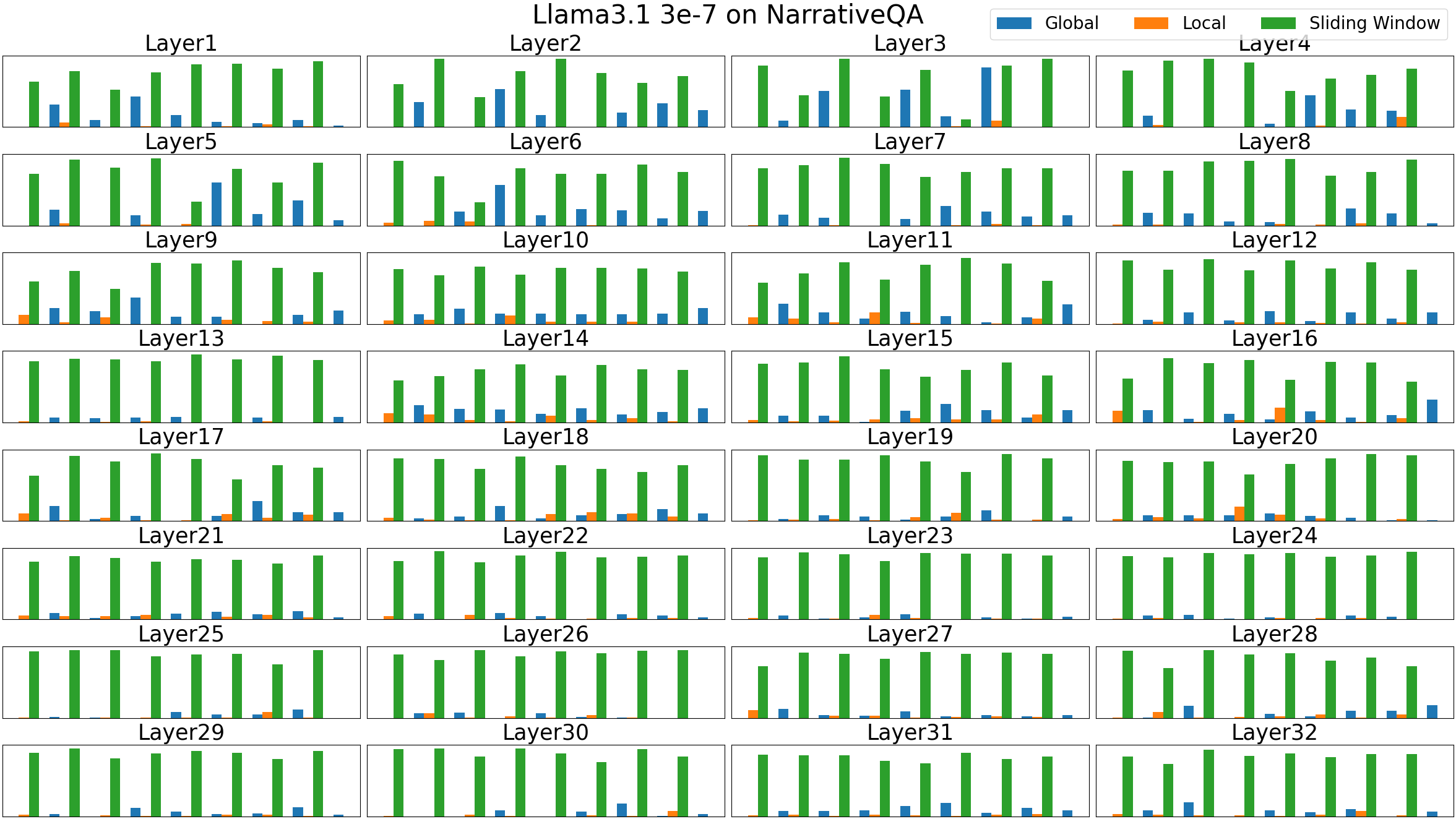}}
    \hfill
    \caption{Number of Token Types for NarrativeQA Dataset~\label{fig:ntokentypenarrativeqa}}
\end{figure}

We also illustrate the token roles distribution in Figure~\ref {fig:ntokentypesmultinew} and~\ref {fig:ntokentypenarrativeqa}for the Multi-News and NarrativeQA datasets. Generally, most tokens are classified as either ~\GlobalTokens{} or ~\SlidingWindowTokens{}, and only a few are considered ~\LocalTokens{}. This might indicate that most attention operations still focus on either long-range or short-term correlations.

\newpage

\end{document}